\documentclass[journal]{IEEEtran}
\usepackage{graphicx,amssymb,mathrsfs,amsmath,array,color,hhline}

\usepackage{textcomp}
\usepackage{stfloats}
\usepackage{url}
\usepackage{verbatim}
\usepackage{cite}
\usepackage{graphicx}
\usepackage{subfigure}
\usepackage{multirow}
\usepackage{booktabs}
\usepackage{algorithm,algorithmic}
\usepackage{cite}
\usepackage{pifont}
\usepackage[colorlinks,urlcolor=blue,driverfallback=dvipdfm]{hyperref}
\usepackage{tabularx}  
\usepackage[table]{xcolor}
\usepackage{array}     

\usepackage{etoolbox}
\makeatletter
\patchcmd{\@makecaption}{\scshape}{}{}{}
\makeatletter
\patchcmd{\@makecaption}{\\}{.\ }{}{}
\makeatother

\begin{document}
\title{DS-Net: A Dynamic Split Network for Edge-Cloud Collaborative Inference}

\author{
        Xinyun Zhang, 
        Li Wang*,~\IEEEmembership{Senior Member,~IEEE,}
        Lei Jin*,
        Lianming Xu, 
        and Zhu Han, ~\IEEEmembership{Fellow,~IEEE,}
\thanks{This work was supported in part by the National Natural Science Foundation of China under Grants 62525106, 62171054, 62201089, and 62201071, in part by China Unicom Research Institute 2025-STHZ-BJYDDX-004.}
\thanks{\textit{*Corresponding author:  Li Wang; Lei Jin}}
\thanks{X. Zhang and L. Wang are with the School of Computer Science (National Pilot Software Engineering School), Beijing University of Posts and Telecommunications, Beijing 100876, China. (e-mail: zhangxinyun@bupt.edu.cn, liwang@bupt.edu.cn)
}
\thanks{Lei Jin and L. Xu are with the School of Electronic Engineering, Beijing University of Posts and Telecommunications, Beijing 100867, China. (e-mail: jinlei@bupt.edu.cn, xulianming@bupt.edu.cn)
}
\thanks{ Zhu Han is with the Department of Electrical and Computer Engineering, University of Houston, Houston, TX 77004 USA, and also with the Department of Computer Science and Engineering, Kyung Hee University, Seoul 446-701, South Korea (e-mail: hanzhu22@gmail.com).
}
}

\markboth{Submission to IEEE, ~Vol.~XX, No.~XX, ~XXXX,~2023}
{Shell \MakeLowercase{\textit{et al.}}: }
\maketitle

\begin{abstract}

Collaborative inference serves as a critical methodology in edge intelligence. Existing collaborative inference methods operate by splitting the deep neural network (DNN) model between edge devices and cloud servers. However, these methods need to load different models for multiple split points when the communication or device changes, thus resulting in long switching times and substantial storage overhead on the edge side. To address this issue, we propose a Dynamic Split Network (DS-Net), which not only enables efficient split point switching but also enhances the collaborative speed and accuracy across various scenarios. Specifically, we design a dynamic switching framework that enables flexible selection among multiple split points within a single model, thereby reducing storage, memory, and switching overhead. To mitigate the feature distribution shifts and task-relevant information loss caused by dynamic splitting and high-ratio compression, we further propose a hierarchical compensation mechanism. Local alignment calibrates the decoded feature distribution at each split point, while task-specific global compensation restores task-relevant information at the global level. In addition, we incorporate sample-adaptive global compensation selection into split-point selection based on sample complexity and runtime resource conditions. Experimental results demonstrate that DS-Net reduces memory overhead by 74.6\% and achieves millisecond-level switching latency. Moreover, under a 10 Mbps bandwidth constraint on the Nano platform, DS-Net outperforms state-of-the-art collaborative methods by achieving 1.2\% higher accuracy on the COCO dataset, reducing latency by 42.5\%, and improving throughput by a factor of 1.8$\times$.


\end{abstract}

\graphicspath{{figures/}}


\section{Introduction}
\IEEEPARstart{E}{d}ge devices are increasingly tasked with supporting real-time perception and decision-making in applications such as Unmanned Aerial Vehicle (UAV) inspection \cite{10095516}, emergency rescue \cite{bhattarai2020embedded}, and augmented reality \cite{10234302}. However, the high computational demands of deep neural networks (DNNs) often lead to unacceptable inference latency or excessive energy consumption on resource-constrained edge devices\cite{8763885}. Offloading all computation to the cloud, on the other hand, introduces significant communication delays due to propagation delay and limited bandwidth\cite{8976180}. To address these challenges, collaborative inference (CI)\cite{8736011, 9052677} has emerged as a promising approach. In CI, the DNN model is split between the edge and the cloud: the edge performs lightweight feature extraction, and the resulting intermediate features are transmitted to the cloud for more computationally intensive processing. By transmitting compact intermediate representations rather than raw data or performing full model execution at the edge, CI effectively reduces end-to-end latency while maintaining high model accuracy. 



\begin{figure}
    \centering
    \includegraphics[width=\linewidth]{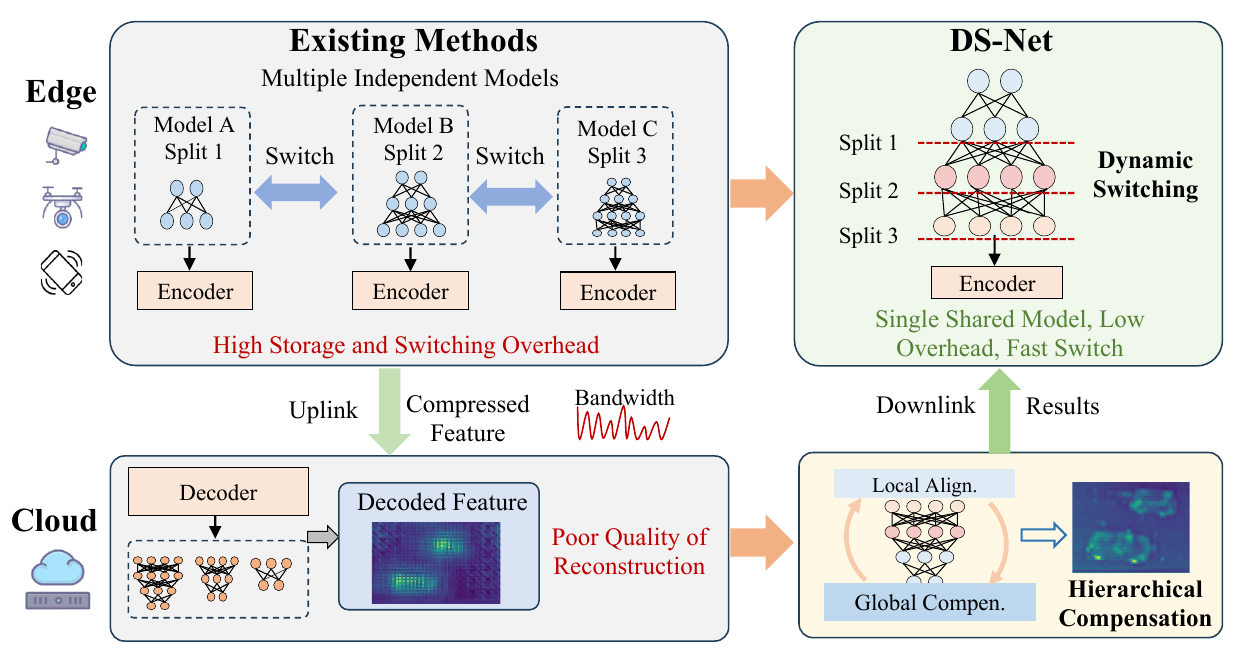}
    \caption{Comparison of existing edge-cloud collaborative methods and DS-Net. Existing approaches require storing multiple model replicas on the edge device, resulting in high memory overhead and model switching latency, while still producing suboptimal feature reconstruction quality on the cloud side. In contrast, DS-Net employs a dynamic switching framework to enable rapid split switching and incorporates a hierarchical compensation module that enhances accuracy under varying bandwidth conditions.}
\label{fig:Three methods}
\end{figure}


In CI, the optimal split point is generally determined by jointly considering the computational capacity of edge devices and the communication bandwidth between the edge and the cloud. The selection of split points generally follows layer-wise \cite{kang2017neurosurgeon, electronics14071345} or graph-based \cite{ eshratifar2019jointdnn} strategies. Furthermore, certain approaches compress intermediate output features after model splitting\cite{ko2018edge,  eshratifar2019bottlenet}. In practical edge-cloud deployments, runtime conditions are often dynamic. Fluctuations in bandwidth may arise from wireless channel variations, device mobility, or network contention, while the available edge computational capacity can vary due to workloads or energy. Because the latency-optimal split point depends on edge computation, feature transmission, and cloud computation jointly, a fixed split point chosen for a predefined scenario may become suboptimal under changing conditions. Thus, dynamic split-point switching is essential to sustain an efficient latency-accuracy trade-off during continuous inference. However, frequent split-point switching also introduces new challenges. As illustrated in Fig. \ref{fig:Three methods}, current methods typically require training and deploying independent models for each possible split point. This approach incurs significant switching latency and memory overhead on edge devices. Moreover, intermediate features at different split points exhibit distinct distributions, making the effects of compression highly split-dependent. Therefore, reducing accuracy degradation across dynamically selected split points while avoiding repeated model switching remains a challenging problem.

To address these limitations, we propose DS-Net, a unified adaptive collaborative inference framework. First, DS-Net integrates multiple candidate split points into a shared backbone and inserts each point with switchable Encoder-Decoder and Identity paths. At runtime, only the Encoder-Decoder path of the selected split point is activated, avoiding the deployment and loading of multiple models. Second, we design a hierarchical compensation mechanism to mitigate feature distortion caused by high-ratio compression. At each split point, local alignment calibrates the decoded feature distribution using the channel-wise mean and standard deviation of the original feature. At the global task level, task-specific compensation extracts spatial information from early-layer features to enhance discriminative and object-related regions. Specifically, saliency and foreground-aware masks are generated for classification and detection, respectively, to recover task-relevant information not captured by local alignment alone. Finally, we incorporate sample-adaptive global compensation selection into the split-point selection process. The split point and compensation state are jointly determined based on sample complexity, real-time bandwidth, and device computing capacity, thereby improving the latency-accuracy trade-off under dynamic runtime conditions.

We conduct extensive experiments on Jetson TX2 and Nano edge devices with an RTX 4090 cloud server. DS-Net is evaluated using four backbone networks and three vision datasets under both controlled and real-world bandwidth conditions. We compare DS-Net with state-of-the-art collaborative inference and lightweight methods in terms of accuracy, latency, throughput, and switching overhead. The results demonstrate that DS-Net enables millisecond-level split-point switching, improves inference efficiency, and effectively alleviates the accuracy degradation caused by model splitting and feature compression.


Our main contributions are summarized as follows:

\begin{itemize}



\item We propose DS-Net, a collaborative inference framework for dynamic edge-cloud environments. DS-Net supports rapid switching among multiple split points within a single model and reduces transmission overhead through feature compression and compensation, while mitigating the accuracy degradation caused by model splitting and high-ratio compression.


\item We propose a hierarchical compensation mechanism to mitigate feature degradation caused by high-ratio compression and decoding. Local alignment calibrates the decoded feature distribution at each split point, while task-specific global compensation enhances task-relevant spatial responses for classification and detection. For detection, global compensation is further enabled adaptively according to sample complexity.


\item Experimental results demonstrate that DS-Net reduces memory overhead by 74.6\% and supports millisecond-level split-point switching. Under a 10 Mbps bandwidth constraint on the Nano platform, DS-Net achieves 1.2\% higher accuracy on the COCO dataset, reduces total latency by 42.5\%, and increases throughput by 1.8$\times$ compared to state-of-the-art collaborative inference methods.
\end{itemize}

In our preliminary conference paper, DMSF\cite{11432620}, we addressed fast split-point switching and multi-branch feature transmission overhead by integrating switchable compression–recovery paths into a unified model and reconstructing FPN multi-scale features from a single transmitted feature. However, the conference version was limited to the YOLOv5 detection task, and structural reconstruction was insufficient to recover task-critical information lost under high-ratio compression. Therefore, in this journal version, we extend DMSF to DS-Net in four aspects: 1) Task: we support both classification and detection across multiple networks and datasets; 2) Model: we consider the different feature requirements of different tasks, we design task-specific global compensation mechanisms to mitigate the information loss caused by dynamic splitting and feature compression; 3) Adaptability: we consider that samples with different complexity levels benefit differently from compensation, we jointly select the split point and compensation state at runtime; and 4) Extensibility: we extend edge–cloud collaborative inference to multi-edge collaboration and introduce a graph-based method for joint split optimization. We also reorganize the paper and provide more comprehensive ablation studies and performance evaluations under non-stationary bandwidth conditions.

The rest of this paper is organized as follows. Section \ref{sec:rela} studies the related work of collaborative inference. Section \ref{sec:moti} introduces the motivation of the paper. Section \ref{sec:proposed} details the architecture of the proposed DS-Net. Section \ref{sec:exp} presents the experimental results and performance evaluations. Finally, Section \ref{sec:clu} draws a conclusion and prospects.

\section{Related Work}
\label{sec:rela}
This section outlines a comprehensive review of the research in collaborative inference.

\noindent\textbf{Model Splitting.}
DNN model splitting focuses on splitting DNNs according to available bandwidth and computational resources, aiming to achieve an optimal trade-off between performance and efficiency\cite{8876870,9155237,10675335}. ANS\cite{zhang2021autodidactic} dynamically learned the optimal split point during inference through a contextual bandit algorithm. Other approaches\cite{8737614, 8871124,10.1145/3397315,11328908} have adopted graph-based modeling techniques. These methods model multi-branch networks as directed acyclic graphs (DAGs) and apply graph optimization algorithms to find split points. Yan et al. \cite{9758628} formulated split point selection as an optimal stopping problem and combined it with model placement strategies. Mu et al. \cite{10227294} achieved fast accuracy estimation via sparsity control and regression-based prediction models. Certain methods \cite{9685432, 11342295} used the reinforcement learning method to optimize the compression rates and split points simultaneously. 

\begin{figure}[t]
    \centering
    \includegraphics[width=\linewidth]{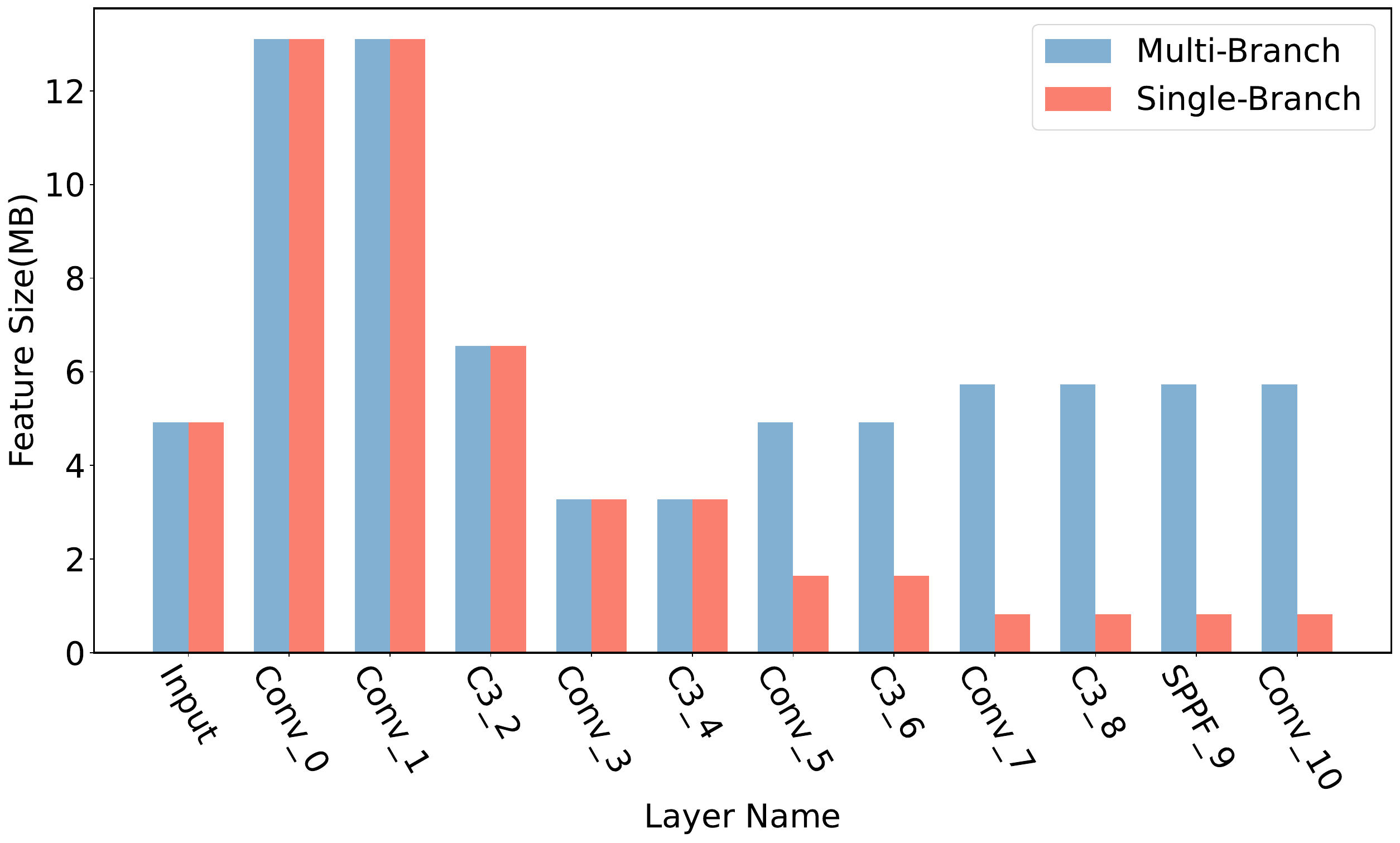}
    \caption{Comparison of feature transmission data size between single-branch and multi-branch of YOLOv5s. SP-Reconfig \cite{9887626} mitigates transmission overhead by using an FRN to transform multi-branch into a single-branch.}
\label{fig:yolov5_model}
\end{figure}

\noindent{\textbf{Feature Compression.}
Existing studies mainly reduced communication overhead in split inference through fixed partitioning and intermediate feature compression \cite{10.1007/978-3-030-01418-6_40, 8451100, 9576096,li2018jalad}. DeepJSCC \cite{9154306} jointly optimized feature pruning and wireless transmission, while Assine et al. \cite{9746043} designed a configurable encoder to trade off bandwidth, computation, and accuracy. AECNN \cite{10279110} and SpikeBottleNet \cite{Hassan2025SpikeBottleNet} compressed intermediate features through channel attention and spiking representations, respectively. Nazir et al. \cite{Nazir2025DistributedSemanticSegmentation} introduced joint source-task decoding for distributed semantic segmentation, directly mapping compressed bottleneck features to segmentation outputs. RCIF \cite{10504739} improved collaborative inference robustness under lossy wireless links, while Azizian et al. \cite{10667003} enhanced feature compressibility by suppressing task-irrelevant information. Wang et al. \cite{10261454} developed DeepAdaIn-Net, which combined high-ratio feature compression, bandwidth-aware configuration, and feature compensation for low-latency AR inference. DVFO \cite{10412103} split intermediate feature maps by attention-estimated importance, preserving critical features while quantizing less important ones to reduce transmission overhead with limited accuracy loss.}

\noindent{\textbf{Joint Splitting-Compression Designs}\label{sec}
In edge-cloud collaborative inference, model splitting and feature compression are closely coupled. The selected split point determines the data size of transmitted intermediate features, while compression directly affects both communication latency and task accuracy. Several studies have jointly considered these two aspects. BottleNet++~\cite{shao2020bottlenet++} introduced autoencoder-based bottleneck units at different split points to reduce feature transmission. Auto-Split~\cite{Auto-Split} combined graph partitioning with mixed-precision quantization under device and accuracy constraints. Shi et al.~\cite{9093772} generated multiple pruned sub-models offline and dynamically selected the optimal configuration online. CNNPC~\cite{9782528} combined channel pruning with 8-bit quantization to search for the optimal split point and compression ratio. SplitNets~\cite{9879512} designed compression modules on the sensor side and recovery modules on the aggregator side, but its complex optimization process potentially compromised training convergence and accuracy at high compression rates. However, many of these methods still depend on split-specific models or compression modules, causing storage and switching overhead in dynamic environments. Moreover, they mainly reduce feature size, but pay limited attention to compression- and reconstruction-induced accuracy degradation across different split points. DS-Net addresses these issues through a unified multi-split model, selected-path compression, and hierarchical feature compensation.}

\section{Motivation}
\label{sec:moti}
In this section, we conduct preliminary experiments to analyze these challenges and motivate the design of DS-Net.

\begin{table}[t]
\centering
\caption{Switching overhead of BottleNet~\cite{shao2020bottlenet++} on the TX2. 
Storage denotes the total model weight size, memory denotes the peak loading memory under the 8192~MB TX2 memory limit, and switching latency denotes the maximum loading time between different split-point models.}
\label{tab:switch_overhead}
\resizebox{0.8\linewidth}{!}{
\begin{tabular}{cccc}
\toprule
Model & Storage (MB) & Memory (MB) & Latency (s)\\
\hline
VGG & 376 & 6,580 & 5.6 \\
ResNet & 139 & 6,836 & 9.6 \\
YOLOv5 & 11.6 & 4,888 & 5.7 \\
YOLOv8 & 8.8 & 5,124 & 3.5 \\
\bottomrule
\end{tabular}}
\end{table}

\begin{table}[t]
\centering
\caption{Performance of FRN with the split point at layer 10. The baseline is the original YOLOv5 multi-branch model. FRN enables single-branch transmission. High Compression applies split-point feature compression, and FRN$\times$1.5 enlarges FRN for accuracy recovery.}
\label{tab:moti_ana}
\resizebox{0.85\linewidth}{!}{
\begin{tabular}{ccc}
\toprule
Method & Feat. (KB)  & Acc.(\%)(COCO/VisDrone) \\
\midrule
Baseline  & 5,200    & 56.5/35.4 \\
FRN  & 400    & 55.5/33.4 \\
FRN + High Compression   & 6  & 53.7/27.2 \\
FRN$\times$1.5            & 6 & 54.1/27.6 \\

\bottomrule
\end{tabular}
}
\end{table}

We first evaluate the switching overhead of existing
split-specific deployment strategies. To evaluate this overhead, we reimplement BottleNet++~\cite{shao2020bottlenet++} on different network architectures and deploy the obtained split-specific models on the TX2 platform. As shown in Table~\ref{tab:switch_overhead}, independent split-point models introduce substantial storage, memory, and switching overhead. In particular, VGG and ResNet incur switching latencies of 5.6 s and 9.6 s, respectively. Although YOLOv5 and YOLOv8 require less storage, their switching latencies are still unsuitable for real-time applications. Since edge computing and UAV perception systems usually require sub-100 ms responsiveness~\cite{9499456}, frequent split-point switching with independent models is impractical. This motivates a unified model design that supports fast split-point adaptation without repeated model loading. 

We then analyze the communication and reconstruction cost in multi-branch vision models. For detectors such as YOLO with feature pyramid structures, splitting at branch layers may require transmitting multiple features, which significantly increases communication latency. As shown in
Fig.~\ref{fig:yolov5_model}, multi-branch feature transmission can be much larger than single-branch transmission. SP-Reconfig~\cite{9887626} alleviates this problem by using a feature reconstruction network (FRN) to reconstruct multi-branch features from a single transmitted feature.

However, reducing the transmitted feature size alone is not sufficient to preserve accuracy. As shown in
Table~\ref{tab:moti_ana}, FRN reduces the transmitted feature size from $5200$ KB to $400$ KB, but still causes accuracy degradation. When high compression is further applied, the feature size is reduced to $6$ KB, while the accuracy drops from $56.5\%$ to $53.7\%$ on COCO and from
$35.4\%$ to $27.2\%$ on VisDrone. Moreover, simply enlarging the FRN brings only limited recovery, indicating that reconstruction capacity alone cannot fully compensate for compression-induced feature distortion. 

Based on these preliminary results, we have the following key observations:
\begin{itemize}
    \item \textbf{High Overhead of Split Point Switching.}
    Since the optimal split point may change with bandwidth and edge-side workload, repeatedly loading different split-specific models causes unacceptable switching latency and memory overhead.
\item \textbf{Accuracy Degradation in Model Splitting and feature compression.}
    High-ratio compression and FRN-based reconstruction
    introduce feature distortion and distribution shift, especially for dense object detection scenarios where fine-grained information are critical.
\end{itemize}
These observations motivate the design of DS-Net. First, we introduce a dynamic switching framework that integrates multiple candidate split points into a single model and activates only the selected encoder-decoder path at runtime, thereby reducing redundant deployment and switching overhead. Second, we introduce a hierarchical compensation module on the cloud side. Local alignment mitigates
split-specific distribution shifts caused by compression and decoding, while global compensation enhances task-related spatial responses after feature reconstruction. These designs collectively enable DS-Net to achieve rapid switching of split points in dynamic edge-cloud environments while minimizing accuracy loss.

\begin{figure*}
    \centering
    \includegraphics[width=\textwidth]{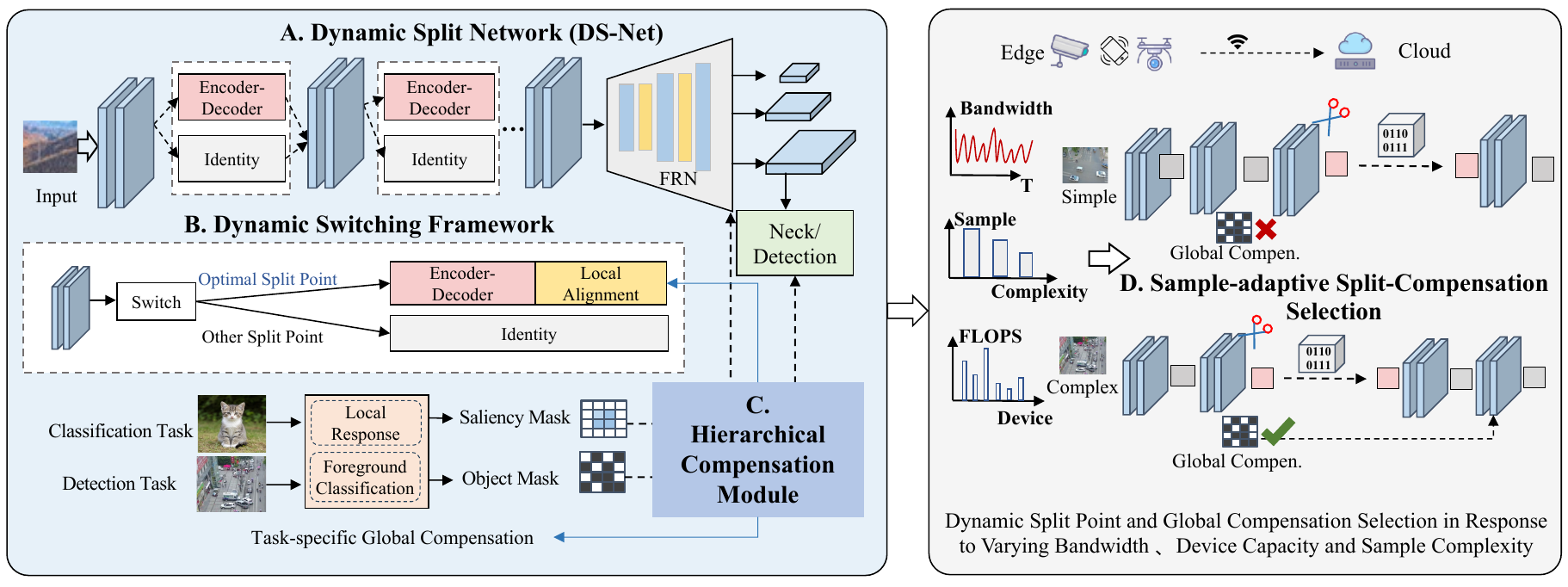}
    \caption{Architecture of DS-Net for edge-cloud collaborative inference. The selected encoder-decoder path enables dynamic split switching, while local alignment and task-specific global compensation mitigate feature degradation. The split point and global compensation state are jointly selected according to bandwidth, device capacity, and sample complexity.}
\label{fig:outline}
\end{figure*}


\section{Dynamic Split Network (DS-Net) for Edge-cloud Collaborative Inference} \label{sec:proposed}
In this section, we present the architectural details of DS-Net. We begin with an overview of DS-Net, followed by a detailed description of the dynamic switching framework. Subsequently, we introduce the hierarchical feature compensation module. Finally, we describe the split point selection and sample-adaptive compensation employed in DS-Net for collaborative inference.

\subsection{Overview of the proposed DS-Net}\label{sec:B}

Fig.~\ref{fig:outline} illustrates the overall architecture of DS-Net. 
DS-Net first introduces a dynamic switching framework, in which each candidate split point is equipped with a switchable encoder-decoder path and an identity path. At runtime, only the encoder-decoder path corresponding to the selected split point is activated to compress and transmit the intermediate feature, while the remaining candidate points forward their features through the identity paths. For FPN-based multi-branch detection networks, DS-Net retains only the main backbone feature for transmission and deploys a feature reconstruction network (FRN) on the cloud side to approximately reconstruct the multi-scale branch features required by the neck. To mitigate the distortion introduced by high-ratio compression, 1-bit quantization, and feature reconstruction, DS-Net further incorporates local alignment and task-specific global compensation. Local feature alignment calibrates the decoded feature distribution using the channel-wise mean and standard deviation of the original split feature, whereas task-specific global compensation enhances task-relevant spatial responses using a saliency mask for classification or a foreground-aware mask for detection. Finally, a sample-adaptive selection mechanism jointly determines the split point and the global compensation state according to input complexity, transmission bandwidth, and device computing capacity.

\subsection{Dynamic Switching Framework}\label{sec:B}

We propose a dynamic switching framework that seamlessly supports both single-branch and multi-branch networks. In this framework, each candidate split point is associated with a dedicated inference path, ensuring that feature compression is performed exactly once at the selected split layer through a switchable encoder–decoder module. This design enables the encoder–decoder modules to deliver targeted, distribution-aware compression for intermediate features at each split point, thereby alleviating feature mismatch and reducing accuracy degradation commonly caused by sharing a single compression module across multiple split points. By integrating all candidate split point models into a unified architecture, the framework allows for fast, runtime selection of split points without redundant parameter loading, significantly minimizing both switching overhead and accuracy loss in dynamic edge–cloud inference scenarios. Next, we provide a detailed description of the switchable encoder-decoder module. Subsequently, we will also describe in detail the feature encoding module at the edge, the decoding module and the reconstruction module in the cloud separately.

\noindent\textbf{Switchable Encoder-Decoder Module.}
To enable dynamic split selection, a switchable encoder–decoder module is integrated after each candidate split point. Each module consists of two parallel pathways: an encoder–decoder path designed for feature compression and recovery, and an identity path that allows features to bypass compression entirely. During runtime, the split point selection mechanism activates only the encoder–decoder path associated with the selected split point, thereby ensuring that compression is applied precisely at that location. For all other (non-selected) split points, the identity branch remains active, passing the original feature maps unchanged. Although all inference paths share the same backbone network, each split point is equipped with its own encoder–decoder module, allowing compression strategies to be tailored to the specific feature characteristics at that depth while maintaining high efficiency and flexibility.


\noindent\textbf{Feature Encoding on the Edge.} 
On the edge side, feature encoding is initially performed using a series of $2\times2$ convolutional layers, which reduce the number of channels in the feature map. This channel-wise compression effectively decreases the data volume for transmission. Let the original feature tensor be $\mathbf{F}_{p} \in \mathbb{R}^{H \times W \times C}$, where $H$, $W$, and $C$ denote the height, width, and number of channels of the feature map, respectively. 
\begin{equation}
\begin{split}
\mathbf{F}_{\text{enc}} =\text{Conv}_{2\times2}(\mathbf{F}_{p}), \quad \mathbf{F}_{\text{enc}} \in \mathbb{R}^{H \times W \times C'},
\end{split}
\end{equation}
where $\mathbf{F}_{\text{enc}}$ denotes the channel compressed feature. 


To further reduce transmission overhead, we employ 1-bit quantization on the channel-compressed features. Specifically, each element of the compressed feature tensor is binarized based on its sign:
\begin{equation}
\operatorname{Sign}(x)=
\begin{cases}
+1, & x \geq 0,\\
-1, & x < 0.
\end{cases}
\end{equation}
The quantized feature is subsequently derived as
\begin{equation}
\mathbf{F}_{\text{bit}} = \operatorname{Sign}(\mathbf{F}_{\text{enc}}), 
\quad 
\mathbf{F}_{\text{bit}} \in \{-1,+1\}^{H \times W \times C'},
\end{equation}
where $\mathbf{F}_{\text{bit}}$ denotes the binarized feature tensor. Compared with the original 32-bit floating-point representation, this binarization reduces the transmitted feature size by up to $32\times$.

\noindent\textbf{Feature Decoding on the Cloud.} 
Feature decoding is performed on the cloud side using a $2\times2$ transposed convolution, which simultaneously restores the original channel dimension and reconstructs the numerical precision of the feature map. This design enables the decoder to recover the channel width $C$ from the compressed $C'$ channels while effectively reintroducing spatial details that may have been attenuated during compression. Since the decoder's convolutional weights are maintained in 32-bit floating-point precision, the 1-bit intermediate features can be directly multiplied by these high-precision parameters. This integration inherently maps binary inputs into continuous-valued activations, thereby eliminating the need for explicit dequantization and facilitating the reconstruction of float32-level feature representations.
\begin{equation}
\begin{split}
\mathbf{F}_{\text{dec}} = \text{DeConv}_{2\times2}(\mathbf{F}_{\text{bit}}), \quad \mathbf{F}_{\text{dec}} \in \mathbb{R}^{H \times W \times C},
\end{split}
\end{equation}
where $\mathbf{F}_{\text{dec}}$ denotes the decoded features. 

\noindent\textbf{Feature Reconstruction on the Cloud.} 
The Feature Reconstruction Network (FRN) draws inspiration from SP-Reconfig~\cite{10279110}, utilizing an upsampling-based architecture to reconstruct low-level features from a single high-level feature map. In DS-Net, the FRN progressively upsamples the received feature representation to restore the spatial resolution required by complex vision tasks. It should be noted that the reconstruction modules in the FRN are not designed as exact mathematical inverses of the corresponding backbone operations. This is due to the inherent non-invertibility of common backbone components, including convolution, C2/C2f/C3 blocks, SPPF, and pooling. Therefore, the FRN is formulated as an upsampling-based reconstruction module, aiming to recover task-relevant feature spatial resolutions rather than exactly reconstructing the original intermediate features.


\begin{table}[t]
\centering
\caption{Exact and approximate operations in the FRN.}
\label{tab:frn_reconstruction}
\resizebox{\linewidth}{!}{%
\begin{tabular}{ccc}
\hline
\textbf{Backbone} & \textbf{FRN Block} & \textbf{Type} \\
\hline
Conv / Downsampling & PixelShuffle / Transposed Conv & Appro. \\
C2 / C2f / C3 & Preserved topology with Transposed Conv & Exact + Appro. \\
BatchNorm & BatchNorm & Exact \\
ReLU / Act. & ReLU / Act. & Exact \\
Concat & Concat & Exact \\
\hline
\end{tabular}%
}
\end{table}

Table~\ref{tab:frn_reconstruction} summarizes the reconstruction design of the FRN. For convolutional operations, PixelShuffle-based upsampling or transposed convolution is used for feature reconstruction. When PixelShuffle is adopted, channel reorganization is used to achieve lightweight sub-pixel upsampling~\cite{Shi_2016_CVPR}. For C2/C2f/C3 blocks, we construct structure-preserving reconstruction modules by replacing the convolutional layers with transposed convolutions while retaining the residual or cross-stage connection patterns. For SPPF and pooling, transposed convolution is used to approximate the recovery of spatial resolution because these operations are inherently non-invertible. BatchNorm, activation functions, and concatenation are retained as standard normalization, nonlinear transformation, and feature-fusion components, respectively, rather than being treated as inverse operators.

Thanks to the FRN module, the feature extraction network only needs to transmit single-layer features from the split point to the cloud, eliminating the need for multi-branch feature transmission and thereby reducing communication overhead. Nevertheless, relying solely on FRN reconstruction may still result in the loss of fine-grained information required by downstream vision tasks. As discussed in the motivation analysis, FRN reduces the transmitted feature size from 5,200 KB to 400 KB, but it still causes accuracy degradation under high compression and splitting, with a drop of 3.8\% on COCO and 8.2\% on VisDrone, as reported in Table~\ref{tab:moti_ana}. This suggests that reconstructing low-level features only from high-level representations is not sufficient for dense small-object scenarios, where object boundaries, local structures, and edge details are particularly important. Moreover, upsampling-based reconstruction may introduce artifacts such as checkerboard patterns~\cite{odena2016deconvolution}, further weakening the spatial fidelity of reconstructed features. Therefore, an additional compensation mechanism is introduced to refine the FRN-reconstructed features and improve their task relevance.


\subsection{Hierarchical Compensation Module}
\label{ssec:comress}
Since dynamic split-point switching involves feature compression and reconstruction at different network depths, different split paths may suffer from different degrees of feature distortion, causing inconsistent accuracy across split points. To alleviate this problem, we propose a hierarchical compensation module consisting of Task-specific Global Compensation and Local Alignment. The task-specific global compensation module enhances task-critical responses in the final shared features by introducing classification- or detection-related spatial masks. In contrast, local alignment mitigates distribution shifts caused by compression and quantization by adjusting the statistical characteristics of the decoded features. By combining these two designs, DS-Net not only reduces compression errors across different split branches, but also enhances task-aware representations for final prediction.

\begin{figure}
    \centering
    \includegraphics[width=1.0\linewidth]{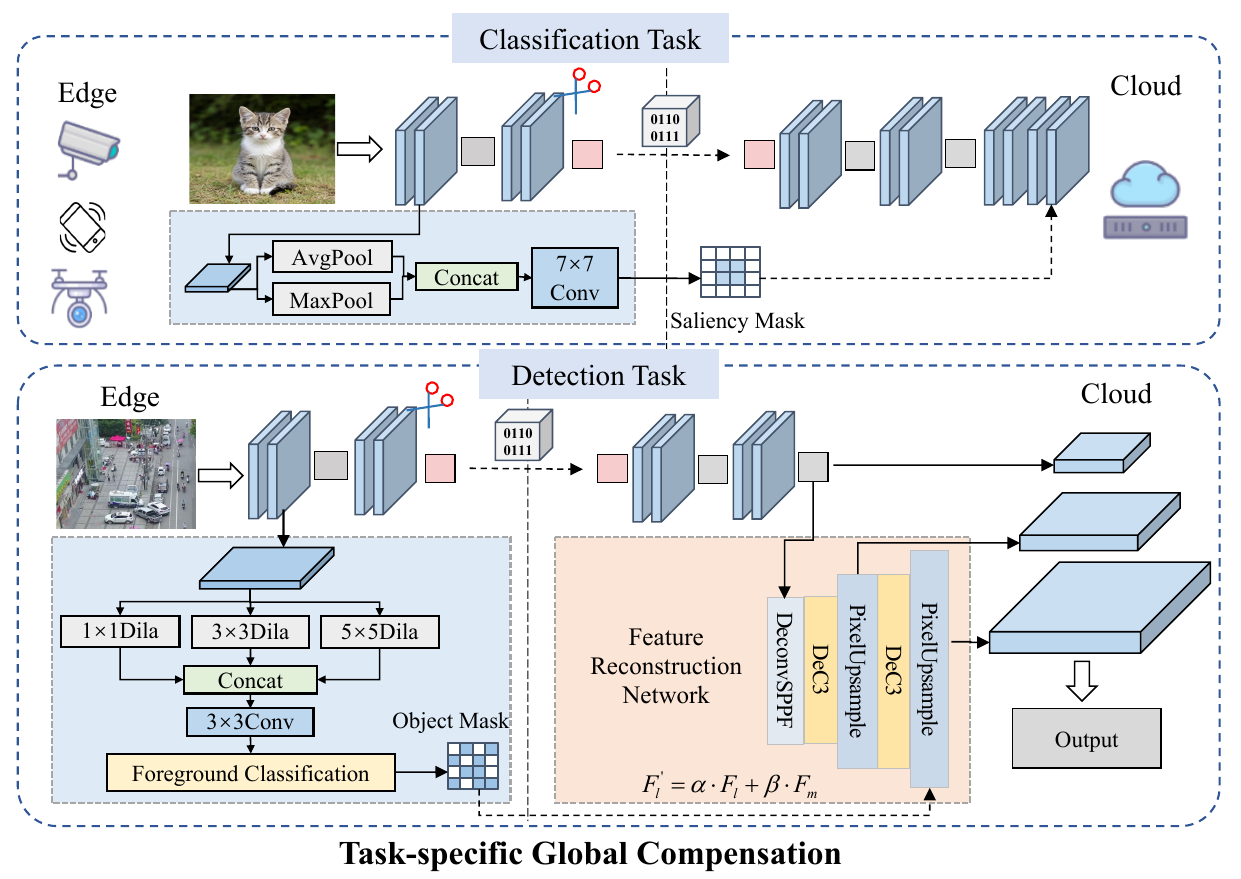}
    \caption{For classification, a saliency mask is generated from early-layer features to enhance discriminative regions in the final shared feature. For detection, an object mask is generated and fused with the FRN-reconstructed features to strengthen object-related spatial responses.}
\label{fig:mask}
\end{figure}

\textbf{Task-specific Global Compensation.}
Fig. \ref{fig:mask} illustrates the detailed information of the task-specific global compensation. The early-layer features preserve richer spatial details and object layout information than deeper features. Therefore, the proposed Task-specific Global Compensation module extracts a single-channel task-related mask from early feature maps. For classification, it learns a saliency response mask to highlight discriminative regions. For detection, it generates an object foreground mask to emphasize object-related spatial responses. The extracted mask is then used to enhance task-critical features in the final shared layer.

For the classification task, the global compensation module generates a saliency mask from early-layer features. Given an early feature map $F_e$, two complementary spatial descriptors are first extracted by channel-wise average pooling and channel-wise max pooling:
\begin{equation}
\mathbf{F}_{\text{avg}} = \operatorname{Pool}_{\text{avg}}(\mathbf{F}_e), \quad
\mathbf{F}_{\text{max}} = \operatorname{Pool}_{\text{max}}(\mathbf{F}_e),
\end{equation}
where $\operatorname{Pool}_{\text{avg}}(\cdot)$ and $\operatorname{Pool}_{\text{max}}(\cdot)$ denote channel-wise average pooling and max pooling, respectively. These two spatial descriptors capture the overall activation distribution and the most discriminative local responses. They are then concatenated and fused by a $7 \times 7$ convolutional layer, followed by a sigmoid function to generate the classification-oriented saliency mask:
\begin{equation}
\mathbf{F}_{m}^{cls}
=
\operatorname{Sigmoid}
\left(
\operatorname{Conv}_{7 \times 7}
\left(
[\mathbf{F}_{\text{avg}}, \mathbf{F}_{\text{max}}]
\right)
\right),
\end{equation}
where $[\cdot]$ denotes channel-wise concatenation and $\mathbf{F}_{m}^{cls}$ is a single-channel spatial mask. The generated mask is transmitted together with the compressed intermediate feature to the cloud.

On the cloud side, the task-related mask is resized by bilinear interpolation to match the spatial resolution of the final shared feature map:
\begin{equation}
\tilde{\mathbf{F}}_{m}^{cls}
=
\operatorname{Interp}
\left(
\mathbf{F}_{m}^{cls}, H_s, W_s
\right),
\end{equation}
where $(H_s, W_s)$ denotes the spatial size of the final shared feature map $\mathbf{F}_s$. Then, the final shared feature is refined under the guidance of the resized mask:
\begin{equation}
\hat{\mathbf{F}}_s
=
\mathbf{F}_s
+
\alpha \cdot \tilde{\mathbf{F}}_{m}^{cls},
\end{equation}
where $\alpha$ is a learnable scaling parameter initialized to zero for stable training. Specifically, the mask emphasizes discriminative spatial regions in the final feature representation, thereby compensating for the loss of task-related responses caused by split inference and feature compression.

For the detection task, we generate a foreground-aware mask to enhance object-related regions. Given an early-layer feature map $\mathbf{F}_2$, parallel dilated convolutions with dilation rates of $d=1$, $2$, and $3$ are applied to capture multi-scale contextual information while preserving spatial resolution:
\begin{equation}
\mathbf{Z}_d = \operatorname{DConv}(\mathbf{F}_2; \mathbf{W}_d, d), \quad d \in \{1, 2, 3\},
\end{equation}
where $\operatorname{DConv}(\cdot)$ denotes a dilated convolution with kernel weights $\mathbf{W}_d$ and dilation rate $d$. These outputs are concatenated and fused by a $3 \times 3$ convolution to produce an integrated feature representation:
\begin{equation}
\mathbf{F}_{\text{fusion}}^{det}
=
\operatorname{Conv}_{3 \times 3}
\left(
[\mathbf{Z}_1, \mathbf{Z}_2, \mathbf{Z}_3]
\right),
\end{equation}
where $[\cdot]$ denotes channel-wise concatenation. This fused feature map is then passed through a lightweight classifier followed by a sigmoid function to predict the foreground-aware mask:
\begin{equation}
\mathbf{F}_{m}^{det}
=
\operatorname{Sigmoid}
\left(
\operatorname{Conv}_{cls}
\left(
\mathbf{F}_{\text{fusion}}^{det}
\right)
\right),
\end{equation}
where $\mathbf{F}_{m}^{det}$ is a single-channel foreground mask. The mask highlights likely object regions and provides task-aware spatial guidance for cloud-side feature compensation.

Due to its single-channel and lightweight nature, the mask introduces minimal transmission overhead and is transmitted along with the compressed intermediate features to the cloud. To accommodate different spatial resolutions in the cloud-side decoder, the mask is resized using bilinear interpolation to match each hierarchical feature map:
\begin{equation}
\tilde{\mathbf{F}}_{m,l}^{det}
=
\operatorname{Interp}
\left(
\mathbf{F}_{m}^{det}, H_l, W_l
\right),
\end{equation}
where $(H_l, W_l)$ denotes the spatial size of the reconstructed feature map $\mathbf{F}_l$ at the $l$-th level. The reconstructed feature map is then refined under the guidance of foreground mask:
\begin{equation}
\hat{\mathbf{F}}_{l}
=
\alpha \cdot \mathbf{F}_{l}
+
\beta \cdot
\tilde{\mathbf{F}}_{m,l}^{det}
\label{equ_global_det}
\end{equation}
where $\alpha$ and $\beta$ are learnable coefficients satisfying $\alpha+\beta=1$. This operation enhances foreground-related spatial responses for the reconstructed feature representation. 

For the classification task, we retain the original training objective and do not introduce an additional mask supervision loss. This is because the saliency response mask is learned implicitly from the image-level classification objective and is only used to provide spatial guidance for global compensation.

For the detection task, the foreground-aware mask should explicitly reflect the spatial distribution of target objects. Therefore, we introduce an additional mask loss to supervise the mask generation module. Specifically, the mask loss is computed as the mean squared error between the predicted mask $\mathbf{F}_{m}^{det}$ and the constructed ground-truth mask $\mathbf{F}_{gt}$:
\begin{equation}
\mathcal{L}_{mask}
=
\operatorname{MSE}
\left(
\mathbf{F}_{m}^{det},
\mathbf{F}_{gt}
\right).
\end{equation}
To construct $\mathbf{F}_{gt}$, we first initialize a zero-valued mask with the same spatial size as the predicted mask. For each annotated bounding box, an elliptical probability region is generated according to the box center, width, and height. The probability reaches its maximum value at the object center and gradually decays toward the boundary, thereby providing soft foreground supervision instead of a hard binary mask. The constructed mask is then normalized to $[0,1]$ for stable training. 

Overall, the global compensation module provides task-aware spatial guidance from early-layer features. For classification, it enhances discriminative saliency regions in the final shared feature, while for detection, it strengthens foreground-related responses in multi-scale feature representations. In this way, the proposed compensation mechanism mitigates the accuracy degradation caused by dynamic split selection, feature compression, and feature reconstruction.

\begin{figure}
    \centering
    \includegraphics[width=1.0\linewidth]{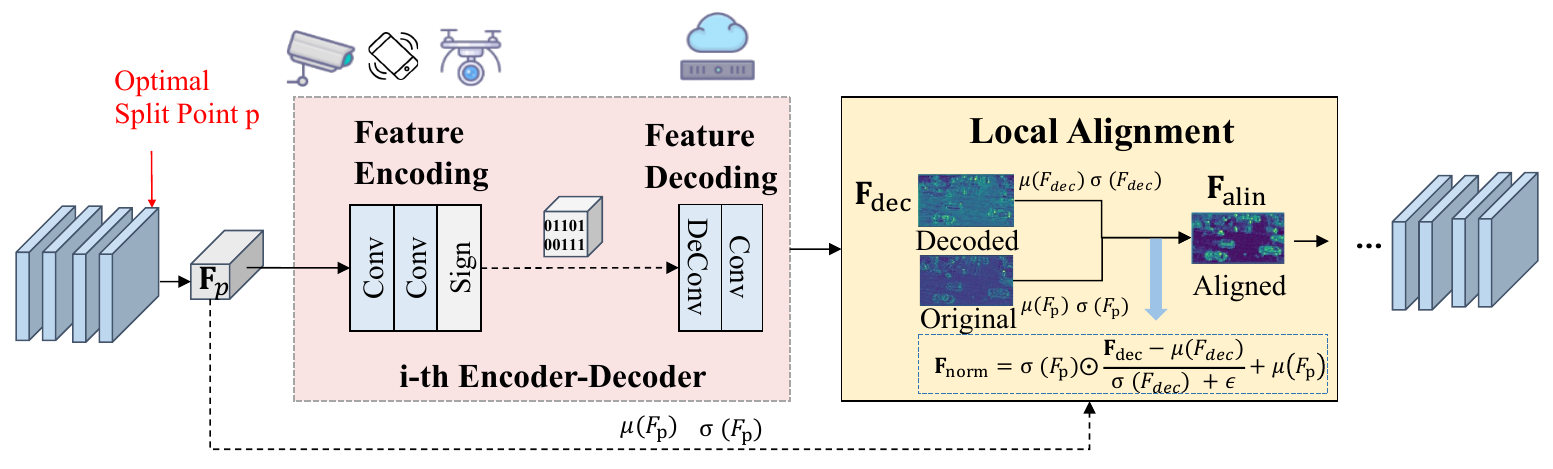}
    \caption{Details of encoder-decoder and local alignment modules. Local alignment achieves feature alignment by transmitting the statistical information of original split features.}
\label{fig:local}
\end{figure}

\noindent\textbf{Local Alignment.} 
We introduce an AdaIN-inspired local feature alignment module~\cite{8237429} to calibrate the decoded features after 1-bit quantization and feature decoding, as shown in Fig. \ref{fig:local}. Different from a generic feature refinement layer, the proposed local alignment is coupled with the quantization-decoding process and operates on the feature decoded from the actual 1-bit compressed representation. Since sign quantization is a many-to-one mapping, activations with different magnitudes but the same polarity are mapped to the same binary value.  As a result, the binary representation mainly preserves the sign pattern, while magnitude, activation energy, and channel-wise scale information are largely discarded. Although the decoder maps the binary representation back to a continuous feature space, the decoded feature may still exhibit channel-wise statistical mismatch with respect to the original split feature. Therefore, we transmit the channel-wise mean and standard deviation of the original split feature as compact side information to calibrate the distributional bias of the decoded feature.

Specifically, the local feature alignment is formulated as
\begin{equation}
\mathbf{F}_{\mathrm{alin}}
=
\sigma(\mathbf{F}_{p}) \odot
\frac{\mathbf{F}_{\mathrm{dec}}-\mu(\mathbf{F}_{\mathrm{dec}})}
{\sigma(\mathbf{F}_{\mathrm{dec}})+\epsilon}
+
\mu(\mathbf{F}_{p}),
\label{equ_adain}
\end{equation}
where $\mathbf{F}_{\mathrm{alin}}\in \mathbb{R}^{H \times W \times C}$ denotes the aligned feature, $\mathbf{F}_{p}$ denotes the original split feature, and $\mathbf{F}_{\mathrm{dec}}$ denotes the decoded feature. The functions $\mu(\cdot)$ and $\sigma(\cdot)$ compute the channel-wise mean and standard deviation, respectively, $\epsilon$ is a small constant for numerical stability, and $\odot$ denotes channel-wise multiplication with broadcasting over the spatial dimensions. 
By matching the decoded feature to the original split feature in terms of channel-wise mean and standard deviation, local feature alignment corrects the activation shift and scale mismatch caused by 1-bit quantization and decoding, thereby mitigating distributional inconsistency during cloud-side inference.

The overall training objective is
\begin{equation}
\mathcal{L}
=
\mathcal{L}_{\mathrm{task}}
+
\lambda_{\mathrm{mask}}\mathcal{L}_{\mathrm{mask}}
\end{equation}
where $\mathcal{L}_{\mathrm{task}}$ is the classification or detection loss, $\mathcal{L}_{\mathrm{mask}}$ supervises the detection foreground mask in global compensation. For the classification task, no additional mask supervision is introduced, which sets $\lambda_{\mathrm{mask}}=0$.
\begin{algorithm}[t]
    \renewcommand{\algorithmicrequire}{\textbf{Input:}}
    \renewcommand{\algorithmicensure}{\textbf{Output:}}
    \caption{Joint Split Point Selection and Sample-adaptive Compensation Inference}
    \label{alg:sample_adaptive_inference}
    \begin{algorithmic}[1]
        \REQUIRE Candidate split-point set $\mathcal{P}$, current
        bandwidth $B_t$, category-specific accuracy profiles
        $\{\mathcal{A}_{b}(p,g)\}$, latency profiles, trade-off
        coefficient $\omega$, and input image $I$.
        
        \ENSURE Selected configuration $(p^{*},g^{*})$ and prediction $\hat{y}$.

        \STATE Generate the compensation mask
        $\mathbf{F}_{m}$.
        
        \STATE Compute the complexity score using
        Eq.~\ref{eq:sample_complexity_score} and determine the sample
        category
        $b\in\{\mathrm{Simple},\mathrm{Medium},\mathrm{Complex}\}$.

        \FORALL{$p\in\mathcal{P}$}
            \FORALL{$g\in\{0,1\}$}
                \STATE Estimate total latency
                $T(p,g)$ using
                Eqs.~\ref{eq:sample_adaptive_latency} and
                \ref{eq:sample_adaptive_transmission}.
                
                \STATE Compute the utility
                $U(p,g)$ using Eq.~\ref{eq:sample_adaptive_utility}.
            \ENDFOR
        \ENDFOR

        \STATE Select $(p^{*},g^{*})$ according to
        Eq.~\ref{eq:sample_adaptive_joint_selection}.

        \STATE Activate the encoder--decoder path corresponding to
        $p^{*}$.

        \STATE Extract the intermediate feature
        $\mathbf{F}_{p}=f_{\mathrm{edge}}^{[1:p^{*}]}(I)$ and compute
        its channel-wise statistics
        $\mu(\mathbf{F}_{p})$ and $\sigma(\mathbf{F}_{p})$.

        \STATE Encode and quantize $\mathbf{F}_{p}$ to obtain
        $\mathbf{F}_{\mathrm{bit}}$.

        \IF{$g^{*}=1$}
            \STATE Transmit $\mathbf{F}_{\mathrm{bit}}$,
            $\mu(\mathbf{F}_{p})$, $\sigma(\mathbf{F}_{p})$, and
            $\mathbf{F}_{m}$ to the cloud.
        \ELSE
            \STATE Transmit $\mathbf{F}_{\mathrm{bit}}$,
            $\mu(\mathbf{F}_{p})$, and $\sigma(\mathbf{F}_{p})$
            to the cloud.
        \ENDIF

        \STATE Decode feature and perform local alignment using Eq.~\ref{equ_adain}.

        \IF{$g^{*}=1$}
            \STATE Apply global compensation using
            Eq.~\ref{equ_global_det}.
        \ENDIF

        \STATE Continue cloud inference and obtain the prediction
        $\hat{y}$.
    \end{algorithmic}
\end{algorithm}

subsection{Split Point Selection and Sample-adaptive Compensation}
\label{sec:sample_adaptive_selection}
Traditional collaborative inference selects split points mainly
according to device computing capacity and transmission bandwidth,
without considering that different inputs may exhibit different
sensitivity to feature compression and compensation. Our analysis
shows that the accuracy gain of global compensation varies with the
foreground complexity of input images. Therefore, based on the dynamic
split framework, we further propose a sample-adaptive mechanism that
jointly selects the split point and global compensation state according
to both input characteristics and current network conditions.


\begin{table}[t]
    \centering
    \caption{Accuracy gains of global compensation across different split points and sample categories on COCO and VisDrone datasets.}
    \label{tab:global_compensation_gain_datasets}
    \setlength{\tabcolsep}{3.8pt}
    \renewcommand{\arraystretch}{1.12}
    \resizebox{\linewidth}{!}{
    \begin{tabular}{c ccc ccc}
        \toprule
        \multirow{2}{*}{Split}
        & \multicolumn{3}{c}{VisDrone}
        & \multicolumn{3}{c}{COCO} \\
        \cmidrule(lr){2-4}
        \cmidrule(lr){5-7}
        & Simple & Medium & Complex
        & Simple & Medium & Complex \\
        \midrule
        Split 0 & +1.3\% & +1.9\% & +2.7\%
                & +0.2\% & +0.3\% & +0.4\% \\
        Split 1 & +0.6\% & +1.6\% & +2.3\%
                & +0.2\% & +0.5\% & +0.7\% \\
        Split 2 & +1.4\% & +2.4\% & +2.5\%
                & +0.5\% & +0.9\% & +0.9\% \\
        Split 3 & +1.6\% & +3.3\% & +4.7\%
                & +0.9\% & +1.2\% & +1.7\% \\
        \midrule
        Average & +1.2\% & +2.3\% & +3.1\%
                & +0.5\% & +0.7\% & +0.9\% \\
        \bottomrule
    \end{tabular}}
\end{table}

\begin{figure}[t]
    \centering
    \subfigure[Distribution of COCOval dataset.]{
        \includegraphics[width=0.46\linewidth]{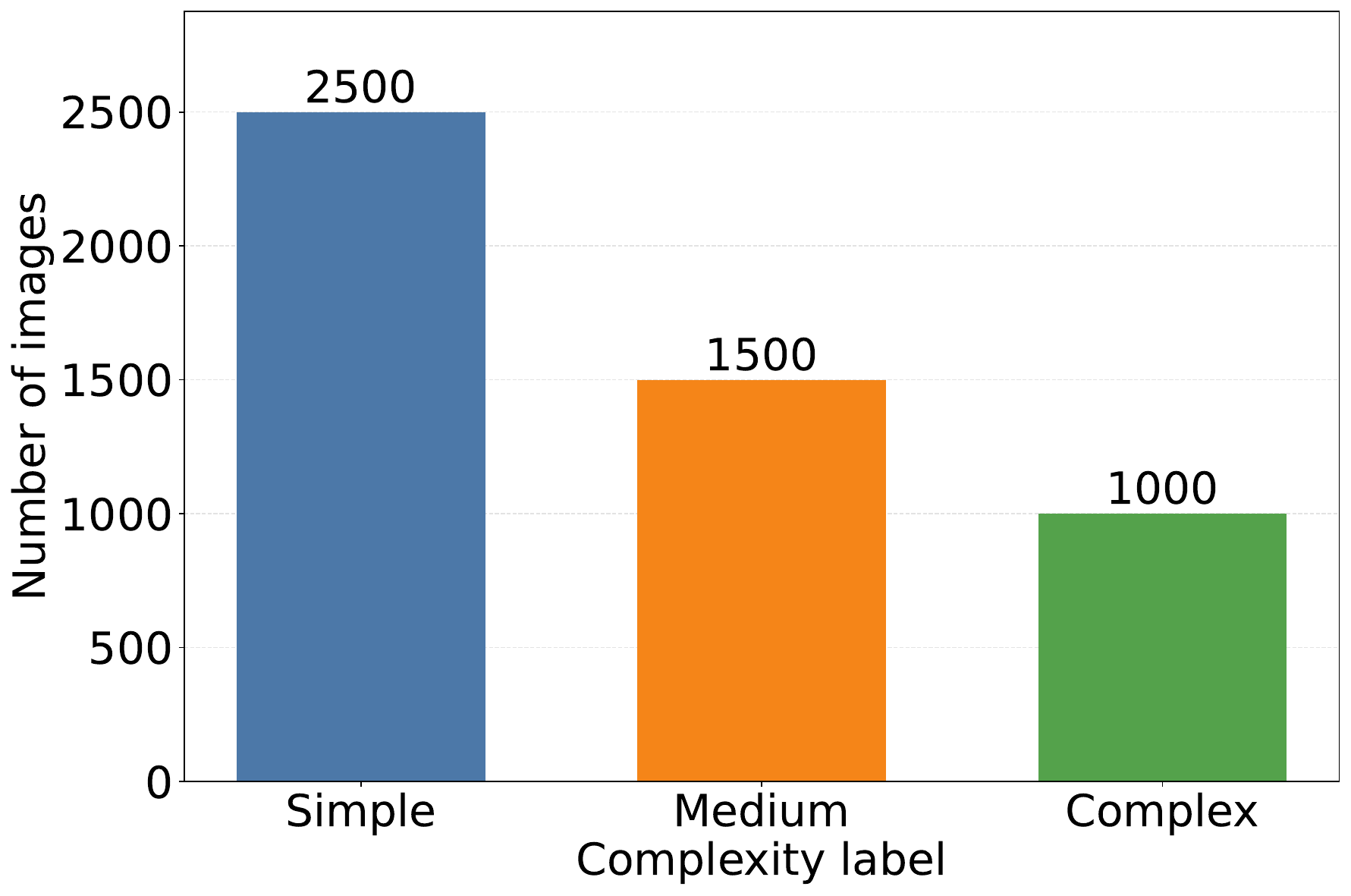}
    }
    \subfigure[Distribution of VisDrone dataset.]{
        \includegraphics[width=0.45\linewidth]{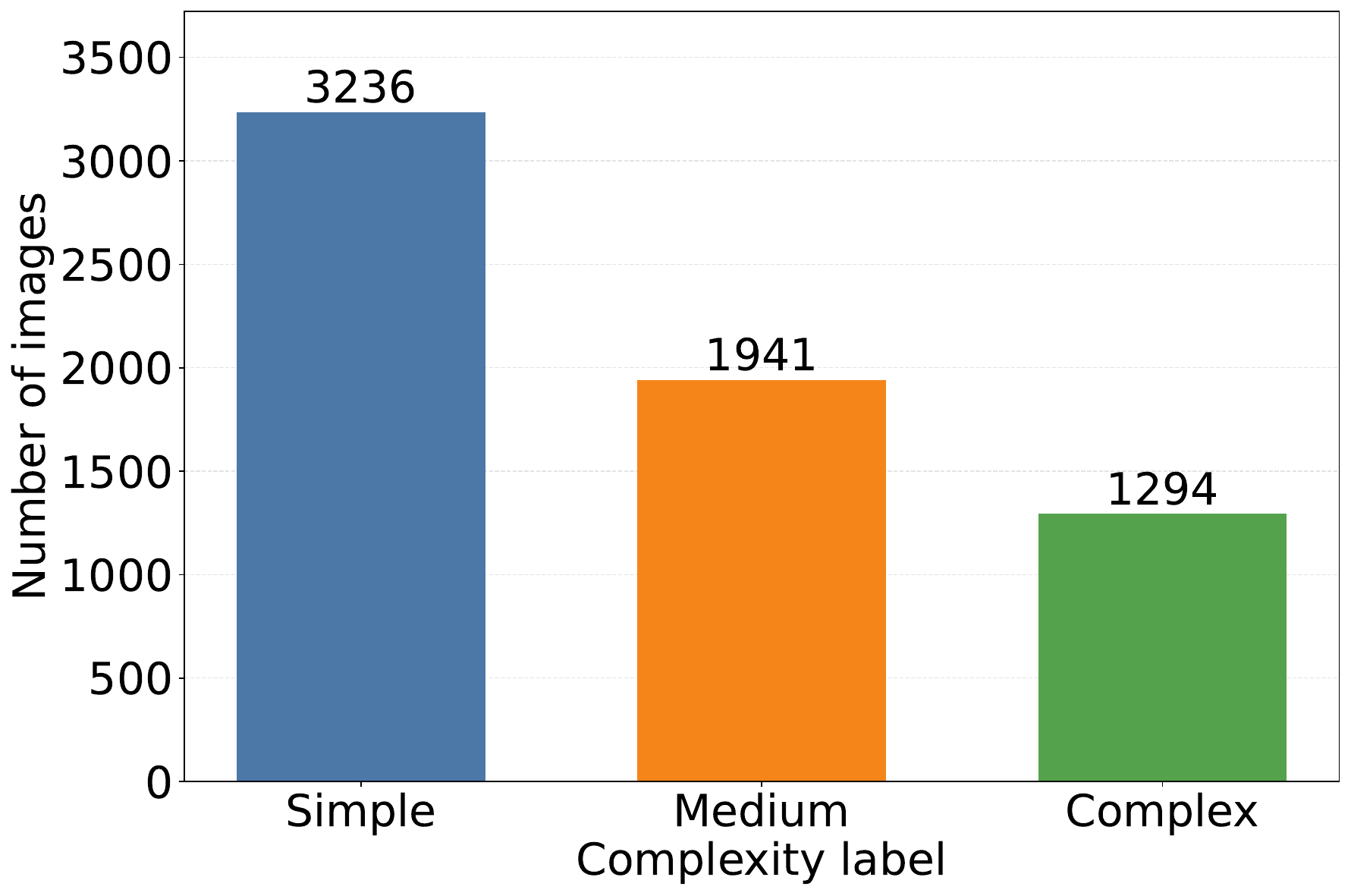}
    }
    \caption{Distributions of foreground-complexity scores on the COCO and VisDrone datasets. The thresholds divide the samples in each dataset into Simple, Medium, and Complex categories.}
    \label{fig:sample_categorization}
\end{figure}

\noindent{\textbf{Sample Categorization for Global Compensation.}
For detection, the global compensation module generates
a foreground-aware mask from early-layer features. Since this mask
reflects the strength and spatial extent of foreground responses, we
reuse it to estimate sample complexity without introducing an
additional prediction network. For each input $x$, the complexity score
is defined as
\begin{equation}
q(x)
=
\frac{1}{H_m W_m}
\sum_{i=1}^{H_m}
\sum_{j=1}^{W_m}
\mathbf{F}_{m}^{det}(x)_{i,j},
\label{eq:sample_complexity_score}
\end{equation}
where $\mathbf{F}_{m}^{det}(x)\in[0,1]^{H_m\times W_m}$ denotes the
predicted foreground-aware mask. A larger $q(x)$ indicates stronger
and more spatially extensive foreground responses. As shown in
Fig.~\ref{fig:sample_categorization}, two dataset-specific thresholds
determined from the profiling set divide the samples into
\emph{Simple}, \emph{Medium}, and \emph{Complex} categories, which
represent relative foreground complexity rather than absolute
detection difficulty. The benefit of global compensation for category
$b$ at split point $p$ is measured by
$\mathcal{A}_{b}(p,1)-\mathcal{A}_{b}(p,0)$. As reported in
Table~\ref{tab:global_compensation_gain_datasets}, this benefit
generally increases with foreground complexity: the average gains for
Simple, Medium, and Complex samples are 1.2\%, 2.3\%, and 3.1\% on
VisDrone, and 0.5\%, 0.7\%, and 0.9\% on COCO, respectively. However,
the sample category only estimates the expected compensation benefit
and does not directly determine the compensation state. All candidate
configurations are retained for online selection.

\noindent{\textbf{Joint Split Point and Global Compensation Selection.}
For an online input $x_t$, its category $b_t$ is first determined from
the foreground-aware mask. Given the current bandwidth $B_t$, the
end-to-end latency of configuration $(p,g)$ is
\begin{equation}
T_t(p,g)
=
T_{\mathrm{edge}}(p,g)
+
T_{\mathrm{trans}}(p,g;B_t)
+
T_{\mathrm{cloud}}(p,g),
\label{eq:sample_adaptive_latency}
\end{equation}
where the transmission latency is
\begin{equation}
T_{\mathrm{trans}}(p,g;B_t)
=
\frac{S_{\mathrm{feat}}(p)+gS_{\mathrm{mask}}}{B_t}.
\label{eq:sample_adaptive_transmission}
\end{equation}
Here, $S_{\mathrm{feat}}(p)$ includes the compressed feature and the
statistics required for local alignment, while $S_{\mathrm{mask}}$
denotes the foreground-mask size.

We define the utility of configuration $(p,g)$ as
\begin{equation}
U_t(p,g)
=
\mathcal{A}_{b_t}(p,g)
-
\omega T_t(p,g),
\label{eq:sample_adaptive_utility}
\end{equation}
where $\omega$ controls the accuracy--latency trade-off. The joint
selection is formulated as
\begin{equation}
\begin{split}
\max_{p,g}\quad
& U_t(p,g) \\
\mathrm{s.t.}\quad
& p\in\mathcal{P},\quad g\in\{0,1\}.
\end{split}
\label{eq:sample_adaptive_joint_selection}
\end{equation}

The selected configuration $(p_t^{*},g_t^{*})$ determines both the
split point and compensation state. Local alignment is always applied
after feature decoding. When $g_t^{*}=1$, the foreground-aware mask is
also transmitted for global compensation; otherwise, mask transmission
and global feature fusion are skipped. Consequently, Simple samples
may favor lower-overhead configurations when compensation provides
limited benefit, whereas Complex samples are more likely to enable
global compensation when its accuracy gain offsets the additional
latency. Algorithm~\ref{alg:sample_adaptive_inference} summarizes the
complete inference procedure.

\begin{table}[t]
\caption{System specifications of edge devices and cloud server.}
\centering
\begin{tabular}{c|ccc}
\toprule
Device & GPU & CPU & Memory \\
\hline
\multirow{2}{*}{Jetson TX2} & 256 cores Pascal & 6-core@2.0GHz & \multirow{2}{*}{8 GB} \\
 & 1.33 TFLOPS& 2$\cdot$Denver2+4$\cdot$A57 & \\
 
\multirow{2}{*}{Jetson Nano} & 128 cores Maxwell & 4-core A57@1.43GHz & \multirow{2}{*}{4 GB} \\
& 472 GFLOPS & Cortex-A57 & \\

\multirow{2}{*}{RTX 4090} & 16384 cores & 36-core@2.10GHz & \multirow{2}{*}{24 GB} \\
 &82.58 TFLOPS & Xeon Gold 6338 & \\
\bottomrule
\end{tabular}
\label{tab:sys}
\end{table}

\section{Experimental Results and Analysis} \label{sec:exp}

\subsection{Experimental Setup}
\noindent\textbf{Prototype.} 
DS-Net employs NVIDIA Jetson TX2 and Jetson Nano as edge devices, and an NVIDIA 4090 GPU as the cloud server. System specifications are summarized in Table \ref{tab:sys}. These devices are interconnected via a wireless network. To emulate varying bandwidth conditions, we utilize the traffic control tool Wondershaper \footnote{https://github.com/magnific0/wondershaper}.

\noindent\textbf{Backbones, Tasks, and Datasets.} 
To evaluate the generality of our framework in vision tasks, we select two representative intelligent applications.
\begin{itemize}
\item \textbf{Image Classification.} We select two widely adopted backbone architectures, VGG-16 \cite{simonyan2014very} and ResNet-50 \cite{He_2016_CVPR}, and evaluate them on the commonly used CIFAR-100 classification dataset\footnote{https://www.cs.toronto.edu/~kriz/cifar.html}.
\item \textbf{Object Detection.} We evaluate YOLOv5s\cite{yolov5} and YOLOv8s \cite{yolov8_ultralytics} on two benchmark datasets: COCO\footnote{https://cocodataset.org/dataset/detection-2017.htm} and VisDrone\footnote{https://github.com/VisDrone/VisDrone-Dataset}, which are widely recognized for object detection evaluation.
\end{itemize}


\noindent\textbf{Comparisons.} 
We compare DS-Net with state-of-the-art methods. \textbf{\emph{(1) Comparison with Collaborative Inference Methods:}} The Edge-only method is executed locally on the edge device as the baseline. The Cloud-only method runs entirely in the cloud, where input images of size $32 \times 32$ for classification and $640 \times 640$ for detection are transmitted from the edge. SP-Reconfig \cite{9887626} applies 1-bit quantization and restores convolutional layers at the split points. DeepAdaIn-Net \cite{10261454} performs adaptive feature compression at the layer immediately preceding the branch structure, using a fixed split point. DVFO \cite{10412103} partitions feature maps according to their importance via an attention mechanism and quantizes the less important features. For each fixed-split baseline, we train or fine-tune the model separately at the corresponding split point with its compression and reconstruction modules enabled. The resulting split-specific checkpoint is used for evaluation, ensuring a fair comparison with DS-Net under optimized configurations. \textbf{\emph{(2) Comparison with Lightweight Models:}} YOLO-nano is a lightweight and compact variant of the YOLO model. Knowledge Distillation (KD) \cite{8953802} is a widely used technique that transfers knowledge from the baseline model to a smaller one. DepGraph \cite{fang2023depgraph} is a dependency graph-based pruning method that removes redundant computations while maintaining model accuracy. YOLO-StarNet \cite{10656434} is an optimized lightweight version of YOLO, enhanced with improved feature extraction mechanisms for more effective object detection.


\noindent\textbf{Evaluation Metrics.}
We evaluate end-to-end latency, throughput, and accuracy. Latency includes edge computation, feature transmission, and cloud computation, while downlink latency is ignored. For pipelined collaborative inference, edge computation and transmission form the first stage, and cloud computation forms the second. Throughput is determined by the slower stage. We report Top-1 accuracy on CIFAR-100 for classification and mAP@0.5 for object detection.




\begin{table}[t]
\caption{Details of DS-Net across various split points, including GFLOPs for the inference model and encoder-decoder module, compressed feature size (Feat.), and accuracy metrics (Top-1 accuracy on CIFAR-100 for ResNet; mAP@0.5 on COCO and VisDrone for YOLOv5).}

\centering
\label{tab:split_res}
\resizebox{\linewidth}{!}{
\begin{tabular}{cccccccc}
\toprule[1.0pt]
\multirow{2}{*}{Model} & Split & \multicolumn{2}{c}{Edge (GFLOPs)} & \multirow{2}{*}{Feat. (KB)} & \multicolumn{2}{c}{Cloud(GFLOPs)} & \multirow{2}{*}{Acc. (\%)} \\
& Layer & Inf. & Enc. & & Dec. & Inf. & \\
\hline
\multirow{4}{*}{ResNet-50}  
& 2 & 0.4 & 0.4 & 16 & 0.4 & 2.0 & 78.0\\
& 3 & 1.1 & 0.4 & 8 & 0.4 & 1.3 & 77.9\\
& 4 & 2.0 & 0.3 & 4 & 0.4 & 0.5 & 77.6\\
& 5 & 2.0 & 0.3 & 2 & 0.4 & 0.0 & 78.0 \\
\hline
\multirow{4}{*}{YOLOv5} 
& 3 & 3.6 & 0.22 & 32 & 0.21 & 28.5 & 55.8/34.3 \\
& 5 & 6.1 & 0.21 & 21 & 0.21 & 26.1 & 56.0/34.5 \\
& 7 & 9.0 & 0.20 & 17 & 0.21 & 23.1 & 56.0/34.5 \\
& 10 & 10.6 & 0.05 & 12 & 0.05 & 21.4 & 56.2/34.2 \\
\bottomrule
\end{tabular}}
\end{table}

\noindent\textbf{Split Point Selection Strategy} 
To achieve flexible and efficient collaborative inference, we select candidate split points at the structural boundaries of the network, specifically at locations where significant changes occur in feature map dimensions or channel numbers. These positions typically correspond to the equilibrium between computational and communication, thereby presenting a higher likelihood of being optimal split points. For ResNet-50, we place four split points at stage boundaries (Conv2\_x–Conv5\_x). In VGG16, we identify the four pooling layers after Conv2–Conv5 as split points. For YOLO-based detectors, we select key backbone layers (3, 5, 7, 10), where significant changes in feature map size and transmission volume occur.

\begin{table*}[htbp]
\centering
\caption{Performance comparison with collaborative inference methods under
10\,Mbps. Acc. denotes Top-1 accuracy on CIFAR-100 for classification and denotes mAP@0.5 on COCO and VisDrone for detection. Lat. represents the total latency, including edge computation, transmission, and cloud computation, while Thpt. denotes the edge-side throughput. Arrows indicate the desired direction of metrics: $\downarrow$ (lower is better), $\uparrow$ (higher is better), and $\rightarrow$ (closer to baseline is better). The Edge-only approach serves as the baseline. We mark the best result as \colorbox{red!15}{red} and the second best one as \colorbox{blue!10}{blue}.}
\label{tab:collaborative_performance_all}

\resizebox{\textwidth}{!}{%
\begin{tabular}{c|c|ccc|ccc}
\toprule
\multirow{2}{*}{Model}
& \multirow{2}{*}{Method}
& \multicolumn{3}{c|}{TX2}
& \multicolumn{3}{c}{Nano} \\
&
& Acc. (\%) $\rightarrow$
& Lat. (ms) $\downarrow$
& Thpt. (img/s) $\uparrow$
& Acc. (\%) $\rightarrow$
& Lat. (ms) $\downarrow$
& Thpt. (img/s) $\uparrow$ \\
\midrule

\multicolumn{8}{c}{\textbf{Classification on CIFAR-100}} \\
\midrule

\multirow{6}{*}{VGG-16}
& Edge-only (base)
& 72.4 & 16.7 & 60
& 72.4 & 14.2 & 70 \\

& Cloud-only
& 72.4 & 83.2 & 12
& 72.4 & 83.2 & 12 \\

& SP-Reconfig \cite{9887626}
& 63.8 & 14.8 & \cellcolor{blue!10}91
& 62.4 & 12.3 & \cellcolor{blue!10}117 \\

& DeepAdaIN-Net \cite{10261454}
& 70.0 & \cellcolor{blue!10}14.3 & 78
& 69.7 & \cellcolor{blue!10}12.0 & 97 \\

& DVFO \cite{10412103}
& \cellcolor{blue!10}71.0 & 16.0 & 66
& \cellcolor{blue!10}71.2 & 13.0 & 85 \\

& DS-Net
& \cellcolor{red!15}71.9
& \cellcolor{red!15}10.7
& \cellcolor{red!15}115
& \cellcolor{red!15}71.5
& \cellcolor{red!15}9.7
& \cellcolor{red!15}130 \\

\midrule

\multirow{6}{*}{ResNet-50}
& Edge-only (base)
& 78.1 & 39.1 & 25
& 78.1 & 46.4 & 22 \\

& Cloud-only
& 78.1 & 87.9 & 11
& 78.1 & 87.9 & 11 \\

& SP-Reconfig \cite{9887626}
& 71.8 & \cellcolor{blue!10}34.5 & 32
& 72.6 & 42.7 & 25 \\

& DeepAdaIN-Net \cite{10261454}
& 74.3 & 34.8 & \cellcolor{blue!10}33
& 74.8 & \cellcolor{blue!10}35.8
& \cellcolor{blue!10}32 \\

& DVFO \cite{10412103}
& \cellcolor{blue!10}77.0 & 35.2 & 29
& \cellcolor{blue!10}77.0 & 38.1 & 27 \\

& DS-Net
& \cellcolor{red!15}78.0
& \cellcolor{red!15}22.9
& \cellcolor{red!15}55
& \cellcolor{red!15}77.9
& \cellcolor{red!15}28.9
& \cellcolor{red!15}42 \\

\midrule
\multicolumn{8}{c}{\textbf{Object Detection on COCO/VisDrone}} \\
\midrule

\multirow{6}{*}{YOLOv5}
& Edge-only (base)
& 56.7/35.4 & 77.2 & 13
& 56.7/35.4 & 192.1 & 5 \\

& Cloud-only
& 56.7/35.4 & 246.2 & 5
& 56.7/35.4 & 246.2 & 5 \\

& SP-Reconfig \cite{9887626}
& \cellcolor{blue!10}55.3/29.6
& 86.5
& 13
& \cellcolor{blue!10}55.3/28.6
& 88.5
& 11 \\

& DeepAdaIN-Net \cite{10261454}
& 48.0/25.7
& \cellcolor{blue!10}63.3
& \cellcolor{blue!10}18
& 48.0/25.7
& \cellcolor{red!15}81.3
& \cellcolor{red!15}14 \\

& DVFO \cite{10412103}
& \cellcolor{blue!10}54.6/32.1
& 115.34
& 10
& \cellcolor{blue!10}54.6/32.1
& 143.7
& 8 \\

& DS-Net
& \cellcolor{red!15}56.0/34.5
& \cellcolor{red!15}61.2
& \cellcolor{red!15}20
& \cellcolor{red!15}55.8/34.3
& \cellcolor{blue!10}82.6
& \cellcolor{red!15}14 \\

\midrule

\multirow{6}{*}{YOLOv8}
& Edge-only (base)
& 61.0/41.1 & 105.4 & 9
& 61.0/41.1 & 261.6 & 4 \\

& Cloud-only
& 61.0/41.1 & 240.3 & 4
& 61.0/41.1 & 240.3 & 4 \\

& SP-Reconfig \cite{9887626}
& \cellcolor{blue!10}59.5/35.3
& 135.6
& 3
& \cellcolor{blue!10}59.6/35.3
& 204.5
& 5 \\

& DeepAdaIN-Net \cite{10261454}
& 57.4/33.2
& \cellcolor{blue!10}81.4
& \cellcolor{blue!10}13
& 57.4/33.2
& \cellcolor{blue!10}174.6
& \cellcolor{blue!10}6 \\

& DVFO \cite{10412103}
& \cellcolor{blue!10}58.0/37.5
& 135.7
& 5
& \cellcolor{blue!10}58.0/37.5
& 235.1
& 4 \\

& DS-Net
& \cellcolor{red!15}60.1/39.6
& \cellcolor{red!15}74.0
& \cellcolor{red!15}14
& \cellcolor{red!15}59.7/39.4
& \cellcolor{red!15}157.7
& \cellcolor{red!15}7 \\

\bottomrule
\end{tabular}%
}
\end{table*}

\noindent\textbf{Implementation and Training Details.} For classification experiments, all models are implemented in PyTorch and trained in the cloud using the SGD optimizer. The initial learning rate is set to $0.1$, the batch size is $128$, the momentum is set to $0.9$, and the weight decay is $5\times10^{-4}$. The models are trained for $200$ epochs using a multi-step learning rate schedule, where the learning rate is multiplied by $0.2$ at epochs $60$, $120$, and $160$. The input images are resized to $32 \times 32$. For object detection experiments, all models are trained using SGD with an initial learning rate of $0.01$, final learning rate factor of $0.2$, momentum of $0.937$, weight decay of $5\times10^{-4}$, and batch size of $32$. Training is conducted for $500$ epochs following the detector training schedule. The learning rate is updated using a LambdaLR scheduler, with the final learning rate set to $0.2$ of the initial learning rate. The input image size is set to $640 \times 640$. Other detector-specific settings, including anchor configuration, NMS threshold, and data augmentation, follow the original baseline implementation. All inference experiments on Jetson Nano and Jetson TX2 are conducted using FP32 precision.

The training procedure of the DS-Net, we employ a single-path training strategy with per-batch path sampling to ensure comprehensive coverage across all possible split paths. Let $\mathcal{P} = \{p_1, \dots, p_K\}$ denote the set of candidate split points. At each epoch and for every mini-batch, a single split point $p$ is uniformly sampled from $\mathcal{P}$, i.e., $p \sim \mathrm{Uniform}(\mathcal{P})$, and only the corresponding encoder–decoder path is activated. A forward and backward pass is then performed exclusively through this active path. The parameters of the shared backbone network are updated jointly with those of the activated path-specific modules. To enable end-to-end training of the 1-bit quantization in the encoder, we apply a straight-through estimator (STE) during the Sign operation.

Table \ref{tab:split_res} presents the detailed analysis of various split points applied to ResNet and YOLOv5. For ResNet, as the split point position deepens, the intermediate features size decreases from 16 KB to 2 KB, with the accuracy range within 0.4\%. For YOLOv5, the size of the intermediate features reduces from 32 KB to 12 KB, and the encoding and decoding module is around 0.2 GFLOPs, with the accuracy range within 0.4\%. 

\subsection{Performance of the DS-Net}

\begin{figure*}[ht]
\centering

\subfigure[Latency on TX2 device.]{
\begin{minipage}[t]{0.23\textwidth}
\centering
\includegraphics[height=3.2cm,keepaspectratio]{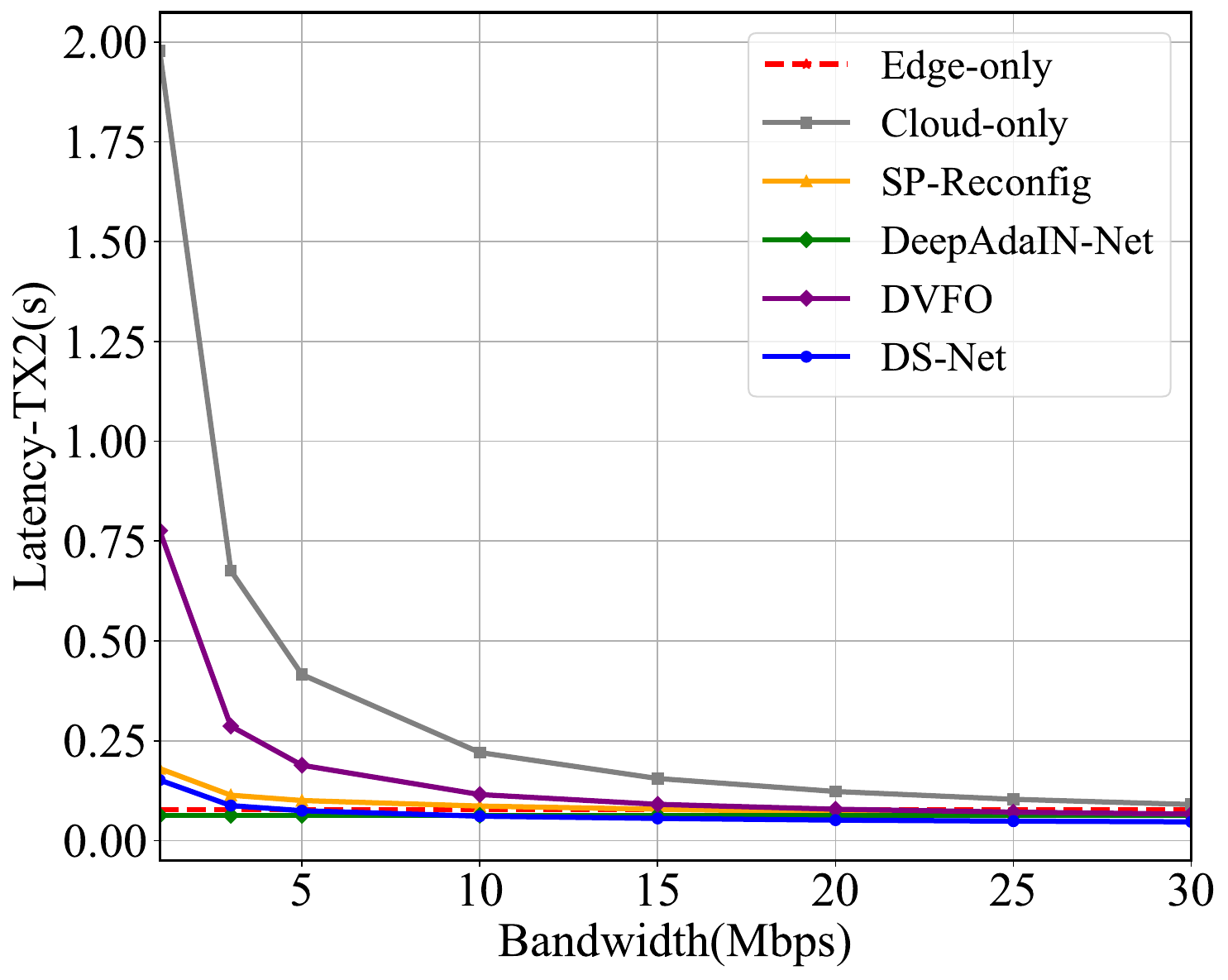}
\label{fig:band_tx2_a}
\end{minipage}%
}
\subfigure[Throughput on TX2 device.]{
\begin{minipage}[t]{0.23\textwidth}
\centering
\includegraphics[height=3.2cm,keepaspectratio]{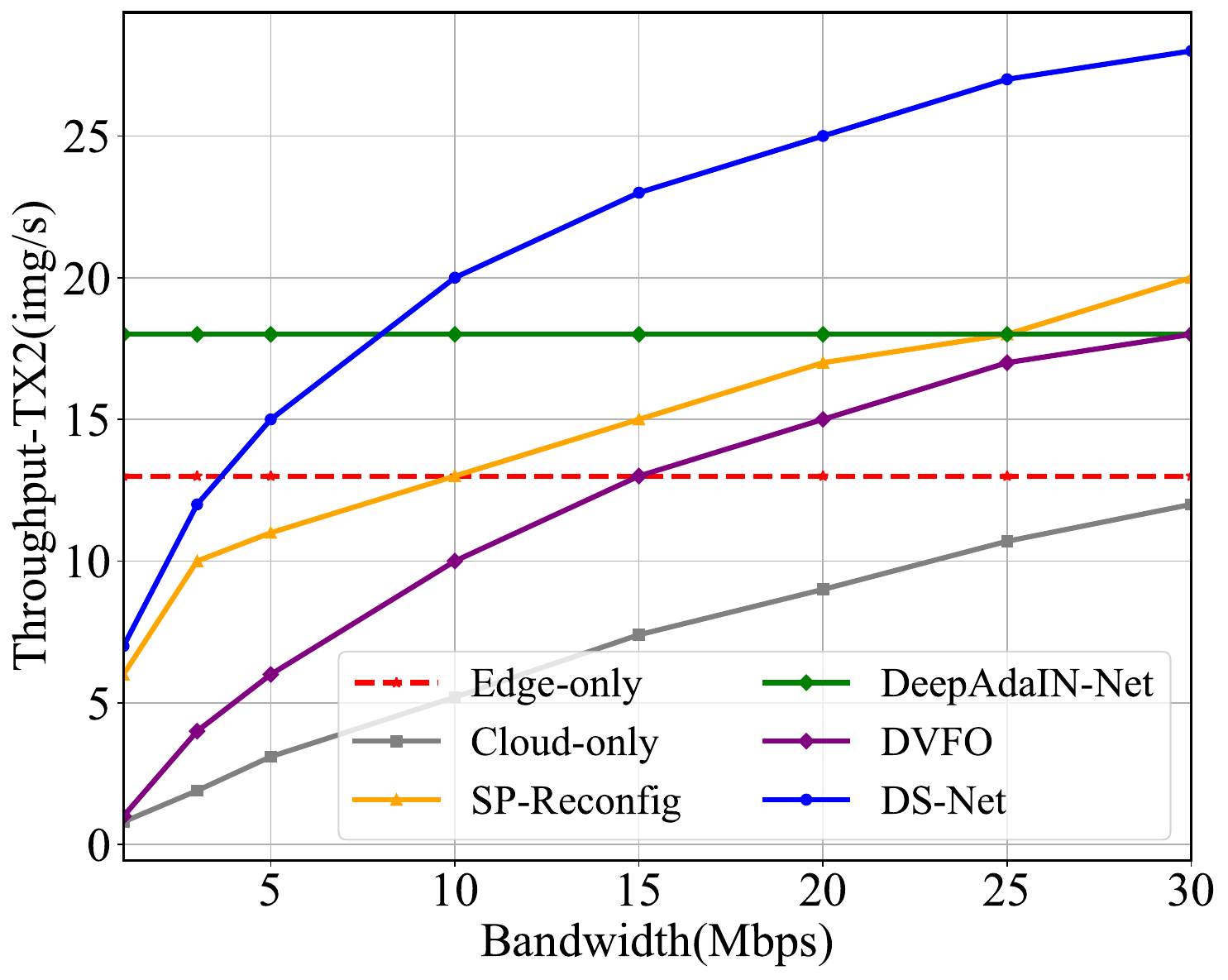}
\label{fig:band_tx2_b}
\end{minipage}%
}
\subfigure[Latency on Nano device.]{
\begin{minipage}[t]{0.23\textwidth}
\centering
\includegraphics[height=3.2cm,keepaspectratio]{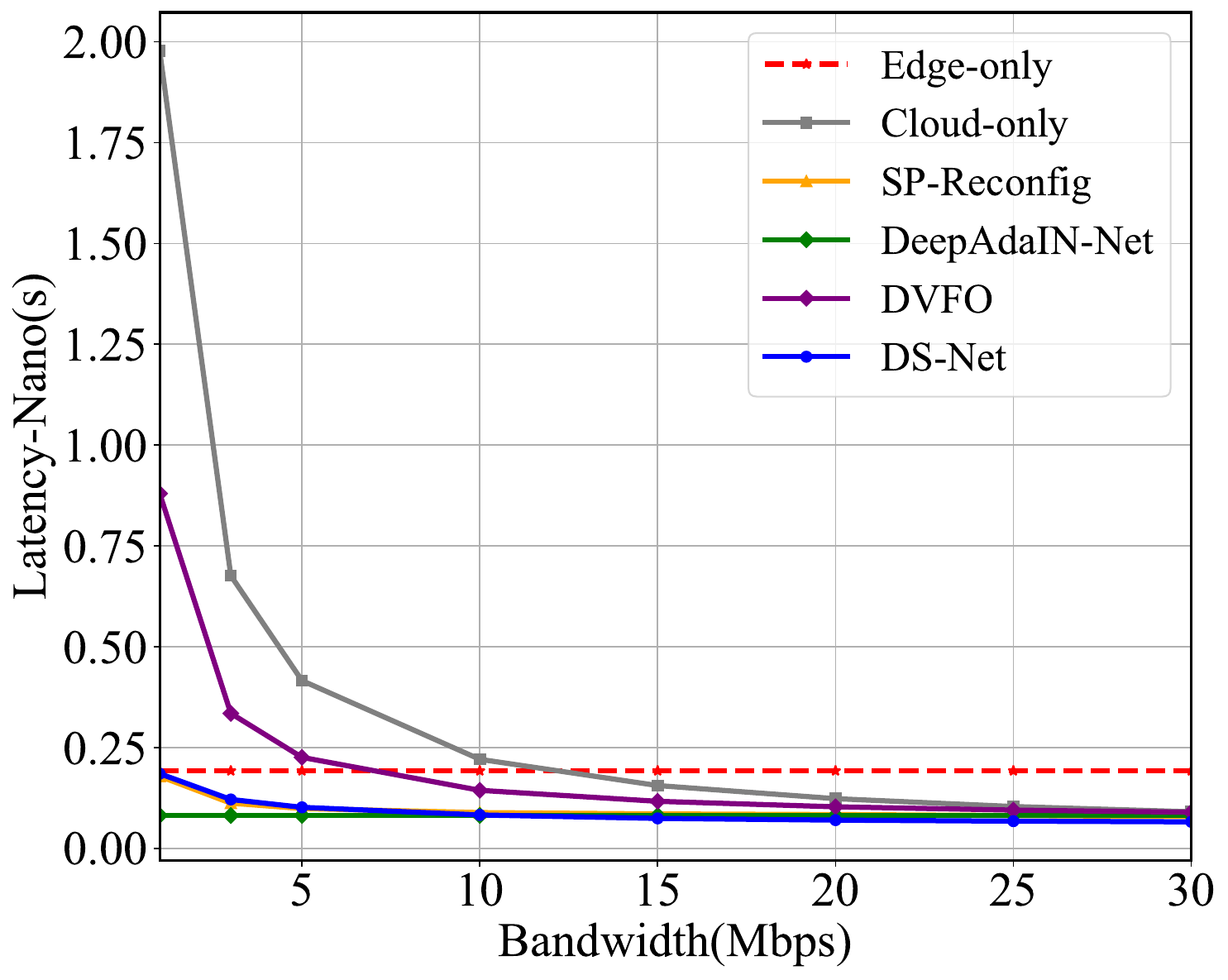}
\label{fig:band_nano_a}
\end{minipage}%
}
\subfigure[Throughput on Nano device.]{
\begin{minipage}[t]{0.23\textwidth}
\centering
\includegraphics[height=3.2cm,keepaspectratio]{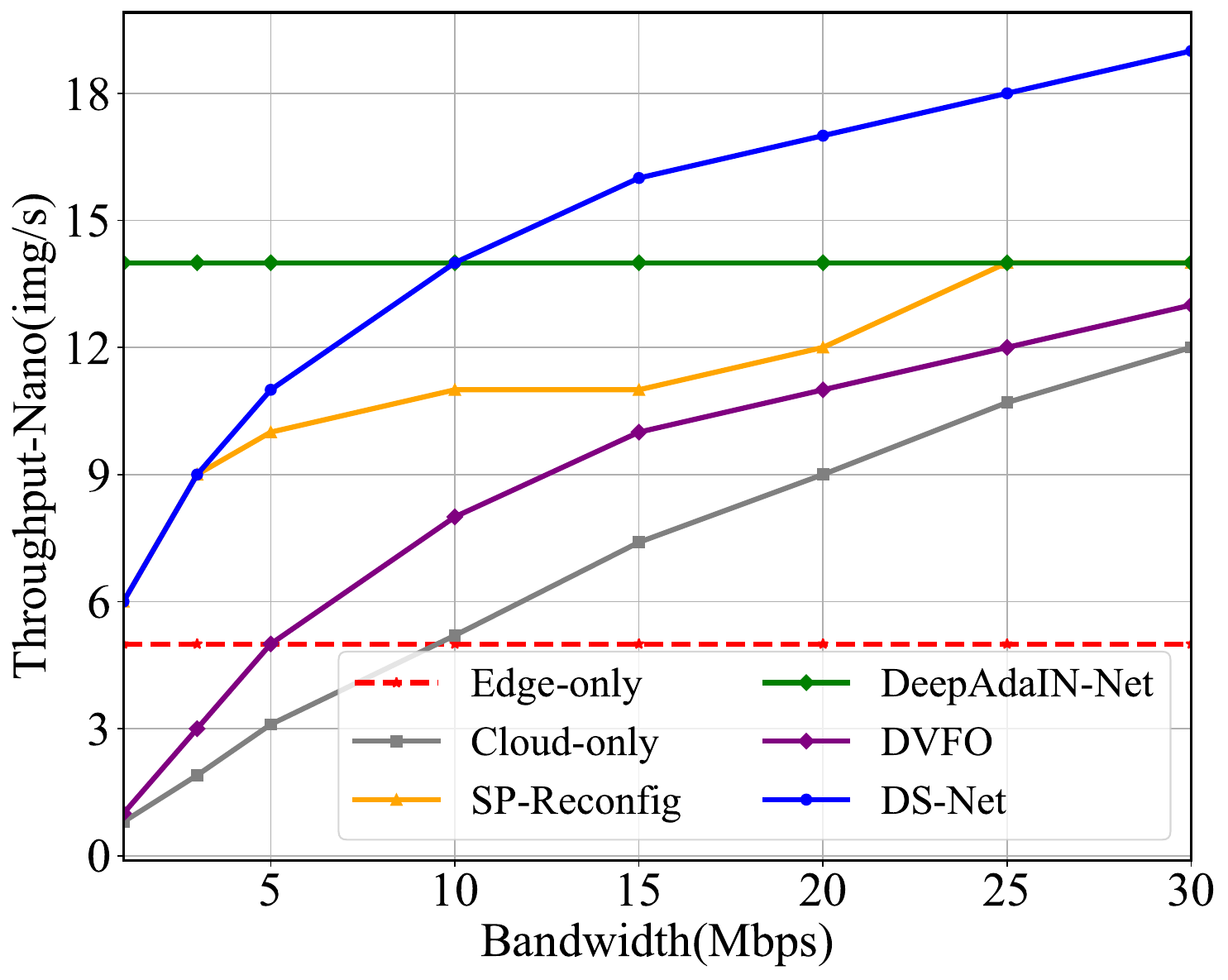}
\label{fig:band_nano_b}
\end{minipage}%
}

\caption{Evaluation of collaborative methods under varying bandwidth conditions on the TX2 and Nano platforms. The accuracy is evaluated using the COCO dataset.}
\label{fig:band_tx2}
\end{figure*}

\begin{figure}[ht]
\centering

\subfigure[Accuracy on COCO dataset.]{
\begin{minipage}[b]{0.48\linewidth}
\label{fig:band_tx2_c}
\includegraphics[height=3.2cm,keepaspectratio]{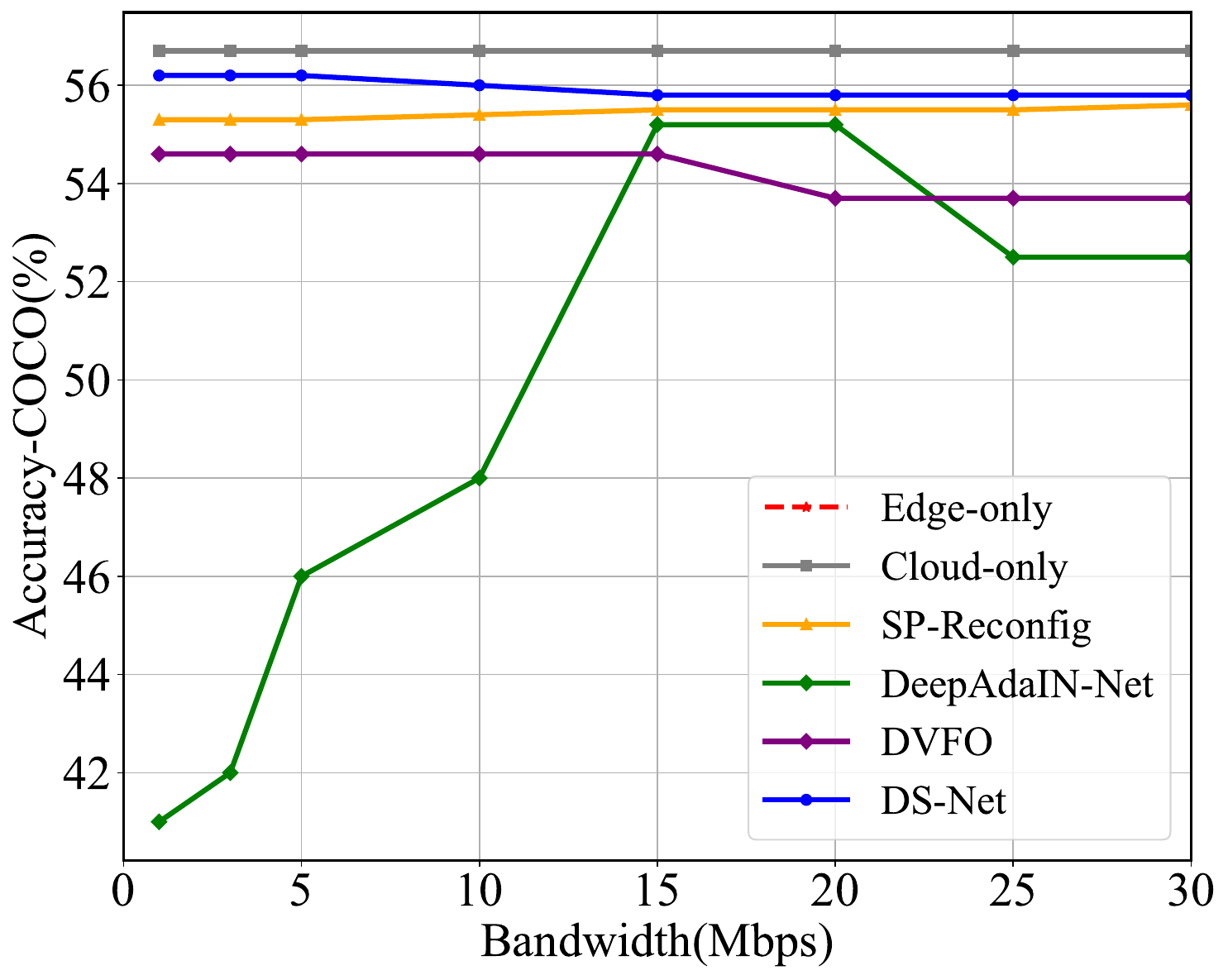}
\end{minipage}%
}
\subfigure[Accuracy on VisDrone dataset.]{
\begin{minipage}[b]{0.48\linewidth}
\includegraphics[height=3.2cm,keepaspectratio]{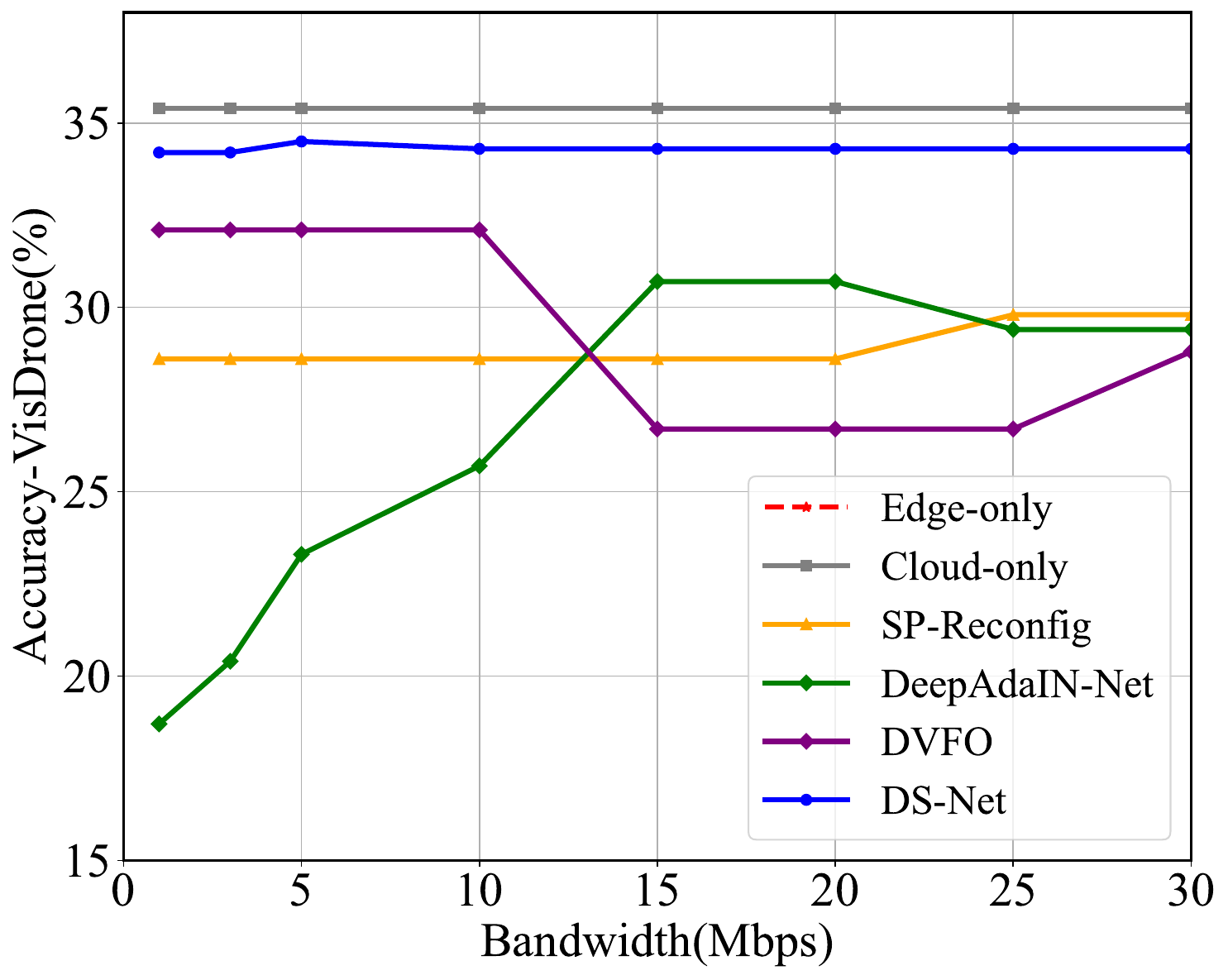}
\label{fig:band_nano_c}
\end{minipage}%
}
\caption{Evaluation of collaborative methods under varying bandwidth conditions on the Nano platform. The accuracy is evaluated using the VisDrone dataset.}    
\label{fig:band_nano}
\end{figure}

\noindent\textbf{Comparison with Collaborative Methods.} 
We evaluate the performance of DS-Net in comparison with state-of-the-art collaborative methods across image classification and object detection tasks. Experiments are conducted on the TX2 and Nano platforms with a controlled bandwidth of 10 Mbps. The results, including inference latency, throughput, and accuracy, are summarized in Tables \ref{tab:collaborative_performance_all}.

\textit{Image classification:} Table \ref{tab:collaborative_performance_all} reports the performance of different collaborative methods on VGG-16 and ResNet-50 under a constrained bandwidth of 10 Mbps. The evaluation was carried out on Jetson TX2 and Jetson Nano. The metrics encompass accuracy on CIFAR, latency, and throughput. The results demonstrate that DS-Net consistently attains the lowest latency and the highest throughput on both TX2 and Nano while maintaining competitive accuracy. Regarding VGG-16, compared to the baseline Edge-only, DS-Net reduces latency by 35.9\% and achieves a 1.9$\times$ throughput speedup on TX2, with only a 0.5\% drop in accuracy. SP-Reconfig outperforms in throughput last time, but at the cost of sacrificing a great deal of accuracy. On TX2, DS-Net outperforms SP-Reconfig with a 1.3$\times$ throughput improvement and an 8.1\% accuracy gain. For the ResNet-50 model, compared to baseline Edge-only on TX2, DS-Net achieves a 2.2$\times$ throughput speedup with only a 0.1\% accuracy drop. DVFO maintains a suboptimal loss in terms of accuracy, but the improvement in latency is relatively low. In contrast, compared with DVFO, the accuracy is improved by 1.0\%, the latency is reduced by 34.9\%, and the throughput is increased by 1.9$\times$. On Nano, compared with DeepAdaIN-Net, the latency has decreased by 19.3\%, the throughput has increased by 1.3$\times$, and the accuracy has improved by 3.1\%. 

\textit{Object detection:} Table \ref{tab:collaborative_performance_all} compares the detection performance of different collaborative methods on YOLOv5 and YOLOv8 under a bandwidth constraint of 10 Mbps, evaluated on Jetson TX2 and Nano. The metrics include accuracy (on COCO and VisDrone), latency, and throughput. For YOLOv5 model, compared to the Edge-only baseline, DS-Net reduces latency by 20.8\% on TX2 and improves throughput by 1.5$\times$ while incurs only a 0.7\% drop on COCO. SP-Reconfig achieves the second-best accuracy. Compared with SP-Reconfig, DS-Net has a 1.5$\times$ speedup and an accuracy improvement of 0.7\% on COCO, while on the VisDrone dataset, its accuracy rate has increased by 4.9\%. Relative to DeepAdaIN-Net, DS-Net attains comparable speed but delivers 8.0\% higher accuracy on COCO and 8.8\% higher accuracy on VisDrone. On Nano, DS-Net achieves 2.8$\times$ improvement in throughput compared to Edge-only baseline, and only a 1.1\% loss on VisDrone in accuracy. DeepAdaIN-Net demonstrates faster speed on the Nano, yet the aggressive compression leads to accuracy loss. Compared with DVFO on Nano, DS-Net achieves a 1.2\% higher accuracy on COCO and 2.2\% higher accuracy on VisDrone, reduces latency by 42.5\%, and increases throughput by 1.8$\times$.

For YOLOv8, DS-Net also outperforms other collaborative methods in the overall trade-off. On TX2, compared to Edge-only baseline, DS-Net reduces latency by 30.0\% and improves throughput by 1.6$\times$. In terms of accuracy, DS-Net incurs only a 0.9\% drop on COCO and 1.5\% on VisDrone datasets. Compared to SP-Reconfig, DS-Net achieves a 4.7$\times$ speedup, while improving accuracy by 0.6\% on COCO and 4.3\% on VisDrone. On Nano, DS-Net maintains competitive accuracy, reduces latency by 39.7\% and achieves 1.8$\times$ improvement in throughput compared to Edge-only baseline. Relative to DeepAdaIN-Net, DS-Net delivers 2.3\% higher accuracy on COCO and 6.2\% higher accuracy on VisDrone. At the same time, it reduces the latency by 9.7\% and increase the speed by 1.2$\times$. Compared with DVFO, DS-Net achieves a 1.9\% higher accuracy on VisDrone, reduces latency by 32.9\%, and increases throughput by 1.8$\times$.

\begin{table*}[htbp]
\centering
\caption{Performance comparison with lightweight methods under 30\,Mbps.
Acc. denotes Top-1 accuracy on CIFAR-100 for classification and denotes mAP@0.5 on COCO and VisDrone for detection. Lat. represents the total latency, including edge computation, transmission, and cloud computation, while Thpt. denotes the edge-side throughput. Arrows indicate the desired direction of metrics: $\downarrow$ (lower is better), $\uparrow$ (higher is better), and $\rightarrow$ (closer to baseline is better). The Edge-only approach serves as the baseline. We mark the best result as \colorbox{red!15}{red} and the second best one as \colorbox{blue!10}{blue}.}
\label{tab:lightweight_performance_all}

\resizebox{\textwidth}{!}{%
\begin{tabular}{c|c|ccc|ccc}
\toprule
\multirow{2}{*}{Model}
& \multirow{2}{*}{Method}
& \multicolumn{3}{c|}{TX2}
& \multicolumn{3}{c}{Nano} \\
&
& Acc. (\%) $\rightarrow$
& Lat. (ms) $\downarrow$
& Thpt. (img/s) $\uparrow$
& Acc. (\%) $\rightarrow$
& Lat. (ms) $\downarrow$
& Thpt. (img/s) $\uparrow$ \\
\midrule

\multicolumn{8}{c}{\textbf{Classification on CIFAR-100}} \\
\midrule

\multirow{4}{*}{VGG-16}
& Edge-only (base)
& 72.4 & 16.7 & 60
& 72.4 & 14.2 & 70 \\

& VGG-11 KD \cite{8953802}
& 69.1
& \cellcolor{blue!10}9.6
& 104
& 69.1
& 11.2
& 89 \\

& DepGraph \cite{fang2023depgraph}
& \cellcolor{blue!10}70.7
& \cellcolor{red!15}8.8
& \cellcolor{blue!10}113
& \cellcolor{blue!10}70.7
& \cellcolor{blue!10}11.0
& \cellcolor{blue!10}91 \\

& DS-Net
& \cellcolor{red!15}71.0
& 10.5
& \cellcolor{red!15}117
& \cellcolor{red!15}71.0
& \cellcolor{red!15}9.6
& \cellcolor{red!15}132 \\

\midrule

\multirow{4}{*}{ResNet-50}
& Edge-only (base)
& 78.1 & 39.1 & 25
& 78.1 & 46.4 & 22 \\

& ResNet-18 KD \cite{8953802}
& 76.9
& 23.0
& 44
& 76.9
& 27.4
& 37 \\

& DepGraph \cite{fang2023depgraph}
& \cellcolor{blue!10}77.4
& \cellcolor{blue!10}20.9
& \cellcolor{blue!10}48
& \cellcolor{blue!10}77.4
& \cellcolor{red!15}21.4
& \cellcolor{blue!10}47 \\

& DS-Net
& \cellcolor{red!15}78.0
& \cellcolor{red!15}18.6
& \cellcolor{red!15}54
& \cellcolor{red!15}78.0
& \cellcolor{blue!10}24.6
& \cellcolor{red!15}50 \\

\midrule
\multicolumn{8}{c}{\textbf{Object Detection on COCO/VisDrone}} \\
\midrule

\multirow{5}{*}{YOLOv5}
& YOLOv5s (base)
& 56.7/35.4
& 77.2
& 13
& 56.7/35.4
& 192.1
& 5 \\

& YOLOv5-nano
& 45.8/25.7
& \cellcolor{red!15}44.2
& \cellcolor{blue!10}23
& 45.8/25.7
& \cellcolor{red!15}66.2
& \cellcolor{blue!10}15 \\

& DepGraph \cite{fang2023depgraph}
& \cellcolor{blue!10}49.5/31.0
& 50.7
& 20
& \cellcolor{blue!10}49.5/31.0
& 74.3
& 13 \\

& YOLOv5-StarNet \cite{10656434}
& \cellcolor{blue!10}51.3/29.8
& 65.0
& 15
& \cellcolor{blue!10}51.3/29.8
& 155.3
& 6 \\

& DS-Net
& \cellcolor{red!15}55.8/34.3
& \cellcolor{blue!10}47.0
& \cellcolor{red!15}28
& \cellcolor{red!15}55.8/34.3
& \cellcolor{blue!10}65.5
& \cellcolor{red!15}19 \\

\midrule

\multirow{5}{*}{YOLOv8}
& YOLOv8s (base)
& 61.0/41.1
& 105.4
& 9
& 61.0/41.1
& 261.6
& 4 \\

& YOLOv8-nano
& 53.1/33.9
& \cellcolor{blue!10}62.4
& \cellcolor{blue!10}16
& 53.1/33.9
& \cellcolor{red!15}138.8
& \cellcolor{blue!10}7 \\

& DepGraph \cite{fang2023depgraph}
& 56.0/37.1
& 80.1
& 13
& 56.0/37.1
& 195.0
& 5 \\

& YOLOv8-StarNet \cite{10656434}
& \cellcolor{blue!10}53.2/34.6
& 88.3
& 12
& \cellcolor{blue!10}53.2/34.6
& 216.3
& 5 \\

& DS-Net
& \cellcolor{red!15}59.7/39.4
& \cellcolor{red!15}56.9
& \cellcolor{red!15}19
& \cellcolor{red!15}59.7/39.4
& \cellcolor{blue!10}140.6
& \cellcolor{red!15}7 \\

\bottomrule
\end{tabular}%
}
\end{table*}

\noindent\textbf{Evaluation under Bandwidth Variation.} 
Fig. \ref{fig:band_tx2} presents the experimental results of various collaborative inference methods under different bandwidth conditions on TX2. As shown in Fig. \ref{fig:band_tx2_a}, the Cloud-only method requires the transmission of a large amount of feature data and is highly sensitive to bandwidth, thus experiencing a significant increase in latency at low bandwidth. The DVFO method also has a relatively large transmission volume and shows high latency when bandwidth is limited. In contrast, although DS-Net has slightly higher latency than Edge-only and DeepAdaIN-Net at extremely low bandwidth, it fully exploits the advantages of split inference as bandwidth increases, and latency drops rapidly. Fig. \ref{fig:band_tx2_b} shows the trend of throughput. With the increase in bandwidth, the throughput of DS-Net continues to improve and achieves the best result among all methods. Due to its dynamic feature compression mechanism, DeepAdaIN-Net has a relatively constant transmission volume, making its throughput insensitive to bandwidth changes. However, its fixed split point limits the overall performance ceiling. SP-Reconfig has a lower throughput than DS-Net. Fig. \ref{fig:band_tx2_c} presents the accuracy results on the COCO dataset. DS-Net has the least accuracy loss under varying bandwidth and is closest to the Edge-only baseline. DeepAdaIN-Net experiences a significant drop in accuracy at low bandwidth, mainly due to the high randomness of its compression strategy leading to information loss. DS-Net, by adaptively adjusting the split position and combining an effective compensation mechanism, effectively maintains stable accuracy performance across different bandwidths, outperforming DVFO and SP-Reconfig.


Fig. \ref{fig:band_nano} presents the experimental results of various collaborative inference methods under different bandwidth conditions on the Nano. As shown in Fig. \ref{fig:band_nano_a}, due to the limited computing power of the Nano, the latency of DS-Net is lower than that of the edge-only baseline at low bandwidth, indicating that the split inference mechanism still has a significant advantage on resource-constrained devices and can achieve efficient collaborative inference with less bandwidth. Fig. \ref{fig:band_nano_b} shows the throughput results, which follow the same overall trend as TX2. DS-Net's throughput continuously increases as the bandwidth gradually increases and ultimately reaches the highest level, demonstrating its excellent scalability even on a more resource-constrained platform. As the bandwidth increases, DS-Net's throughput rate keeps rising and eventually reaches the highest level among all methods, fully demonstrating its good scalability on platforms with limited computing power. In contrast, the edge-only method is constrained by the Nano's computing capacity, and its throughput rate remains at a relatively low level. The throughput rate of DeepAdaIN-Net hardly fluctuates with the change in bandwidth, but its performance improvement at high bandwidth is limited. As shown in Fig. \ref{fig:band_nano_c}, on the VisDrone dataset, DS-Net maintains the smallest accuracy loss compared to other methods in more challenging scenarios, demonstrating strong generalization ability. The accuracy of other collaborative methods fluctuates significantly with the change in bandwidth, among which DVFO has a higher accuracy but insufficient stability. Overall, DS-Net achieves a good balance between accuracy and efficiency under different bandwidth and dataset scenarios, verifying its adaptability and advantages on lightweight devices.

\noindent\textbf{Comparison with Lightweight Models.} 
We evaluate the performance of DS-Net in comparison with lightweight models on image classification and object detection tasks. Experiments are conducted on the TX2 and Nano with a bandwidth of 30 Mbps. The results, including inference latency, throughput and accuracy, are summarized in Tables \ref{tab:lightweight_performance_all}.

\textit{Image classification:}
Table \ref{tab:lightweight_performance_all} evaluates the performance of VGG-16 and ResNet-50 under three lightweight methods, KD, DepGraph, and DS-Net, with a 30 Mbps bandwidth constraint. For VGG-16, both DepGraph and DS-Net significantly reduce FLOPs, with DS-Net achieving the lowest computational cost. On TX2, compared to Edge-only, DS-Net reduces VGG-16 latency by 37.1\% and improves throughput by 2.0$\times$, with only a 1.4\% drop in accuracy. DepGraph achieves the fastest inference time due to its lightweight design, albeit at the expense of some accuracy. On Nano, DS-Net reduces VGG-16 latency by 32.4\% and increases throughput by 1.9$\times$ relative to Edge-only. For ResNet-50 on the TX2 platform, DS-Net reduces latency by 52.4\% and improves throughput by 2.2$\times$ compared to edge-only execution, and by 1.1× relative to DepGraph. Unlike lightweight models that depend on pruning to reduce computational load, DS-Net utilizes strategically placed split points combined with compression and compensation modules to minimize both computation and transmission overhead while maintaining rich feature representation.

\textit{Object detection:} Table \ref{tab:lightweight_performance_all} compares the detection performance of YOLOv5 and YOLOv8 under various lightweight methods, including YOLO-nano, YOLO-s, DepGraph, StarNet, and DS-Net. In terms of computational cost, YOLOv5s and YOLOv8s require 16.5 GFLOPs and 28.8 GFLOPs, respectively, whereas DS-Net requires only 6.3 GFLOPs and 6.8 GFLOPs, reducing computation by approximately 61.8\% and 76.4\%. Although YOLOv5-nano and YOLOv8-nano achieve the lowest FLOPs and parameter counts, they suffer an accuracy drop of around 10\% compared to their baseline counterparts. In terms of accuracy, DepGraph suffers a 7.2\% drop on COCO and a 4.4\% drop on VisDrone, whereas DS-Net incurs only 0.9\% and 1.1\% degradation, respectively. For YOLOv5 on the TX2 platform, DS-Net achieves 1.4$\times$ higher throughput than DepGraph while improving accuracy by 6.3\% on COCO. On the Nano platform, DS-Net reduces YOLOv5’s latency by 65.9\% and increases throughput by 3.8$\times$ compared to the baseline, outperforming DepGraph with 11.8\% lower latency and 1.5$\times$ higher throughput. For YOLOv8 on TX2, DS-Net reduces latency by 46.0\% and doubles throughput relative to the baseline, achieving 35.6\% lower latency and 1.6$\times$ higher throughput compared to StarNet. Regarding accuracy, StarNet experiences a 7.8\% decline on COCO and a 6.5\% decline on VisDrone, while DS-Net maintains significantly better performance with only 1.3\% and 1.7\% accuracy loss.

\begin{figure*}[ht]
\centering

\subfigure[Dynamic uplink bandwidth.]{
\begin{minipage}[b]{0.23\textwidth}
\label{fig:vbandwidth2}
\includegraphics[width=1.0\textwidth]{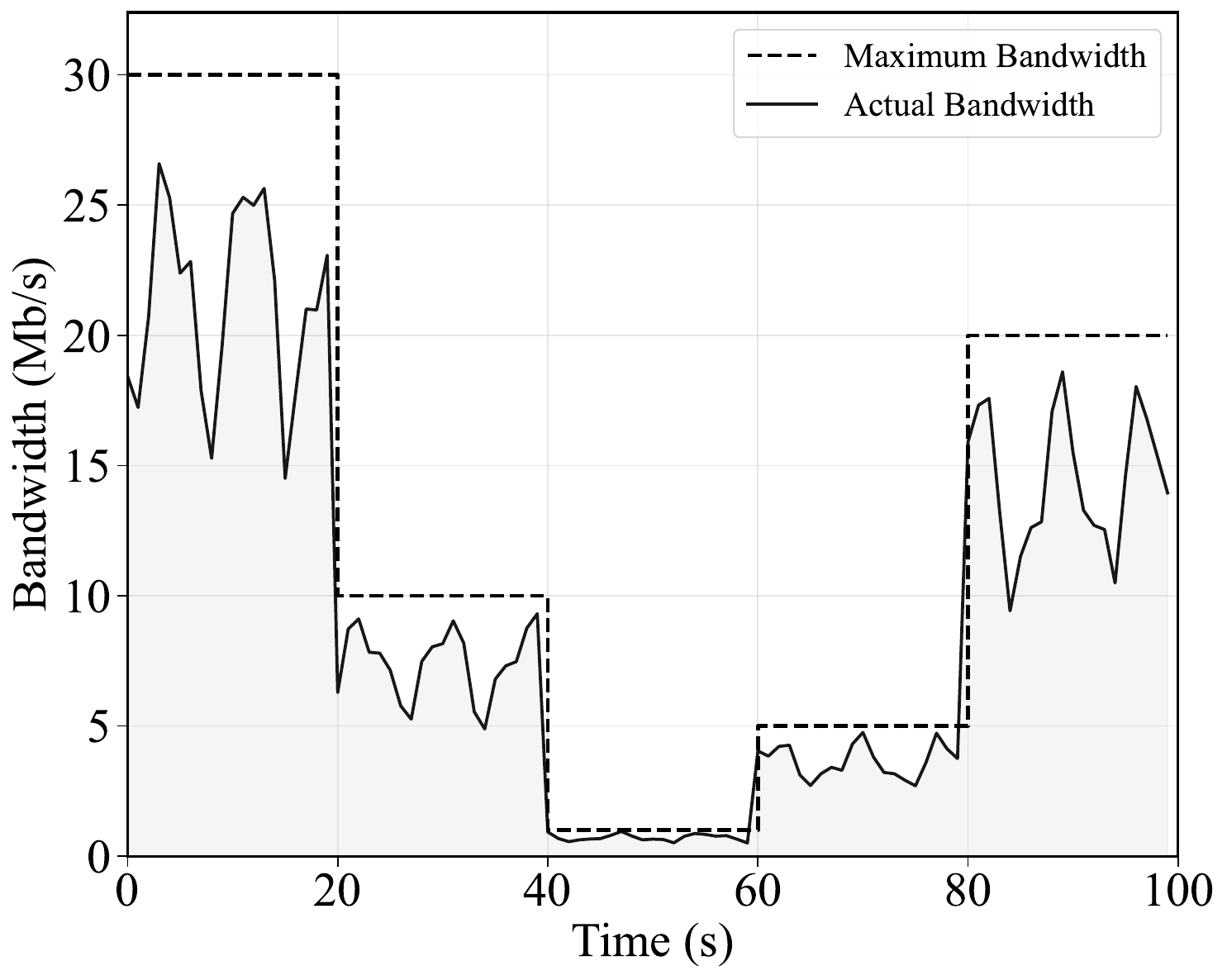}
\end{minipage}%
}
\subfigure[Split-point trajectory of DS-Net.]{
\begin{minipage}[b]{0.23\textwidth}
\label{fig:vbandwidth_split2}
\includegraphics[width=1.0\textwidth]{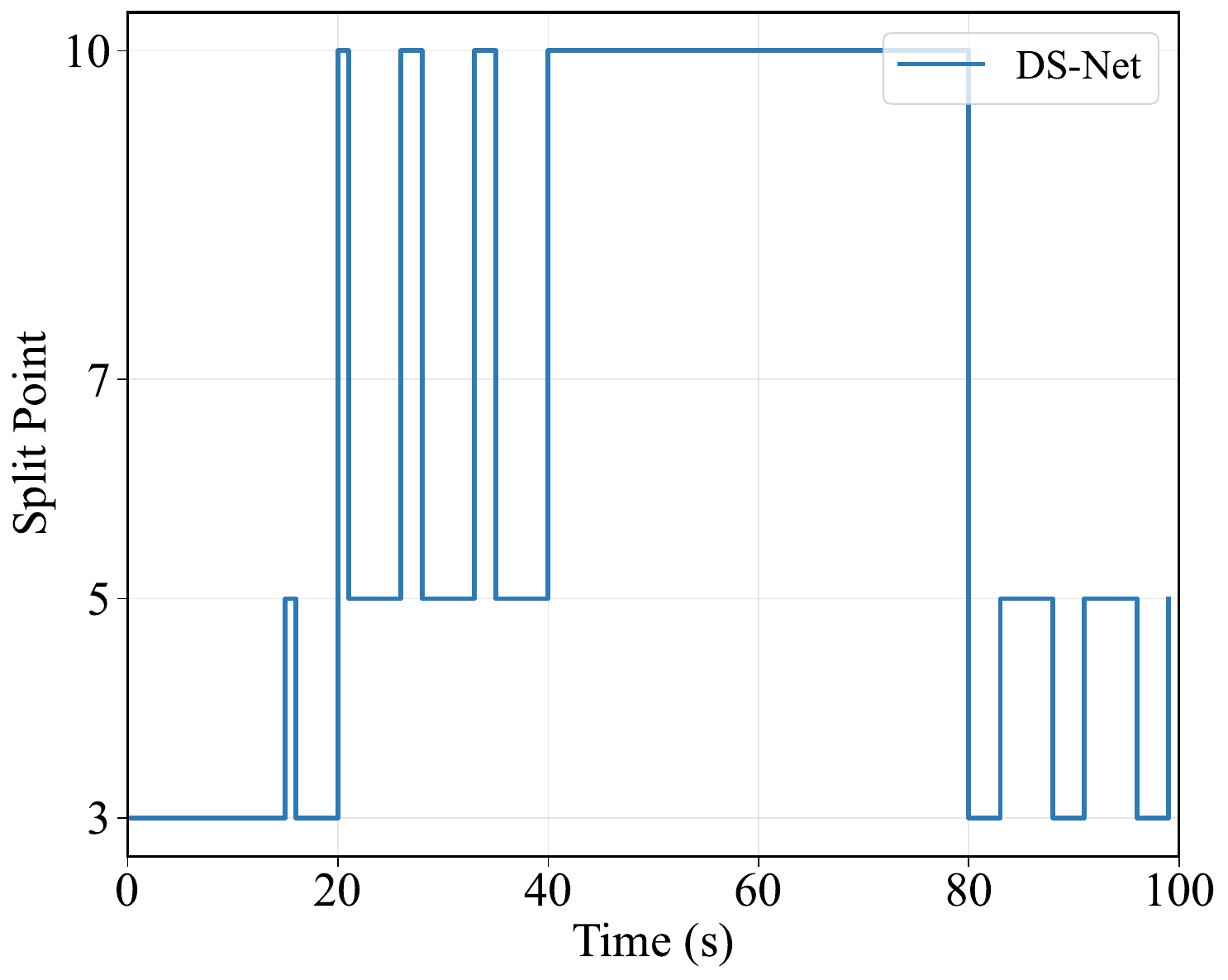}
\end{minipage}%
}
\subfigure[Latency variation of DS-Net.]{
\begin{minipage}[b]{0.23\textwidth}
\label{fig:vbandwidth_lat2}
\includegraphics[width=1.0\textwidth]{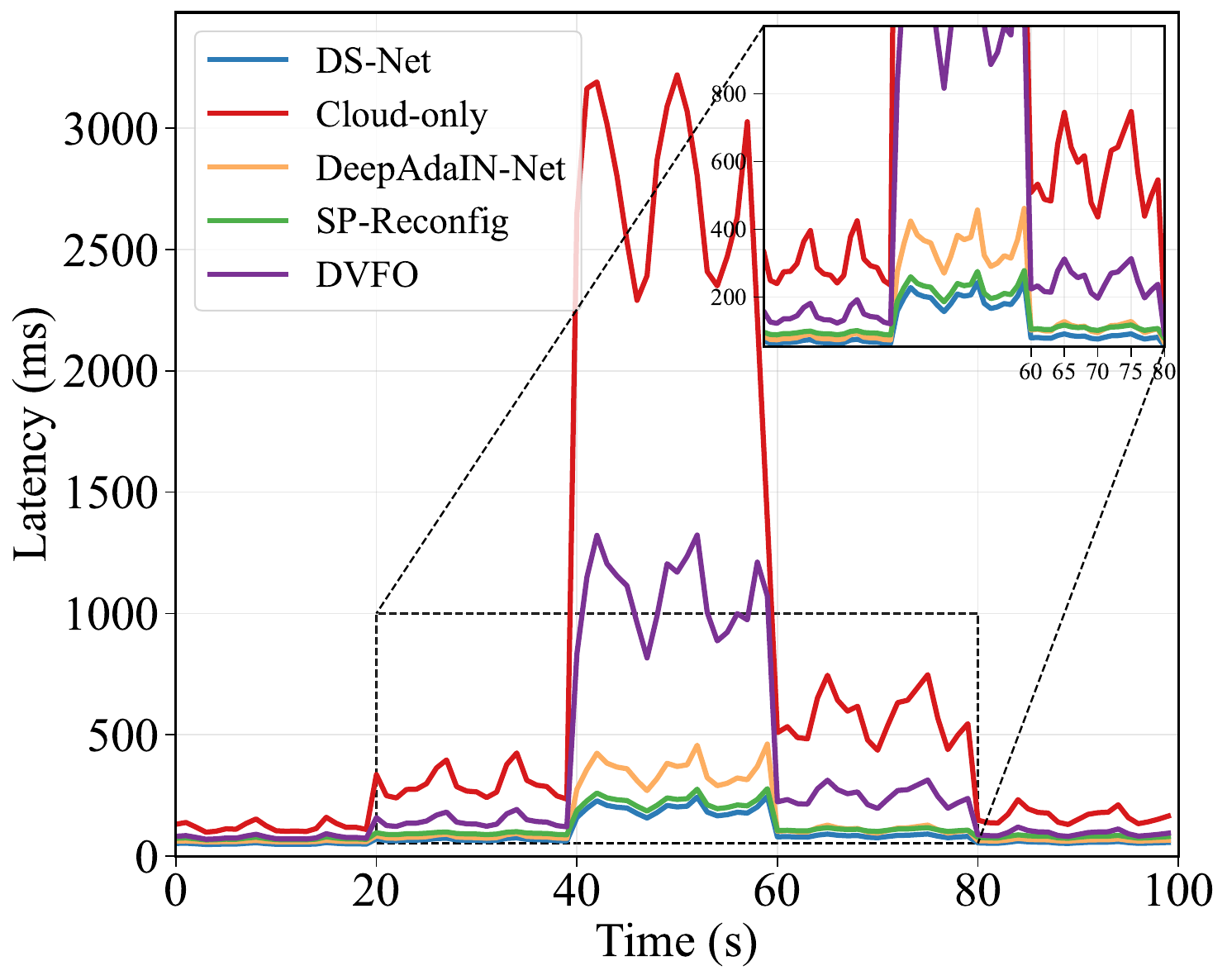}
\end{minipage}%
}
\subfigure[Accuracy variation on COCO.]{
\begin{minipage}[b]{0.23\textwidth}
\label{fig:vbandwidth_acc2}
\includegraphics[width=1.0\textwidth]{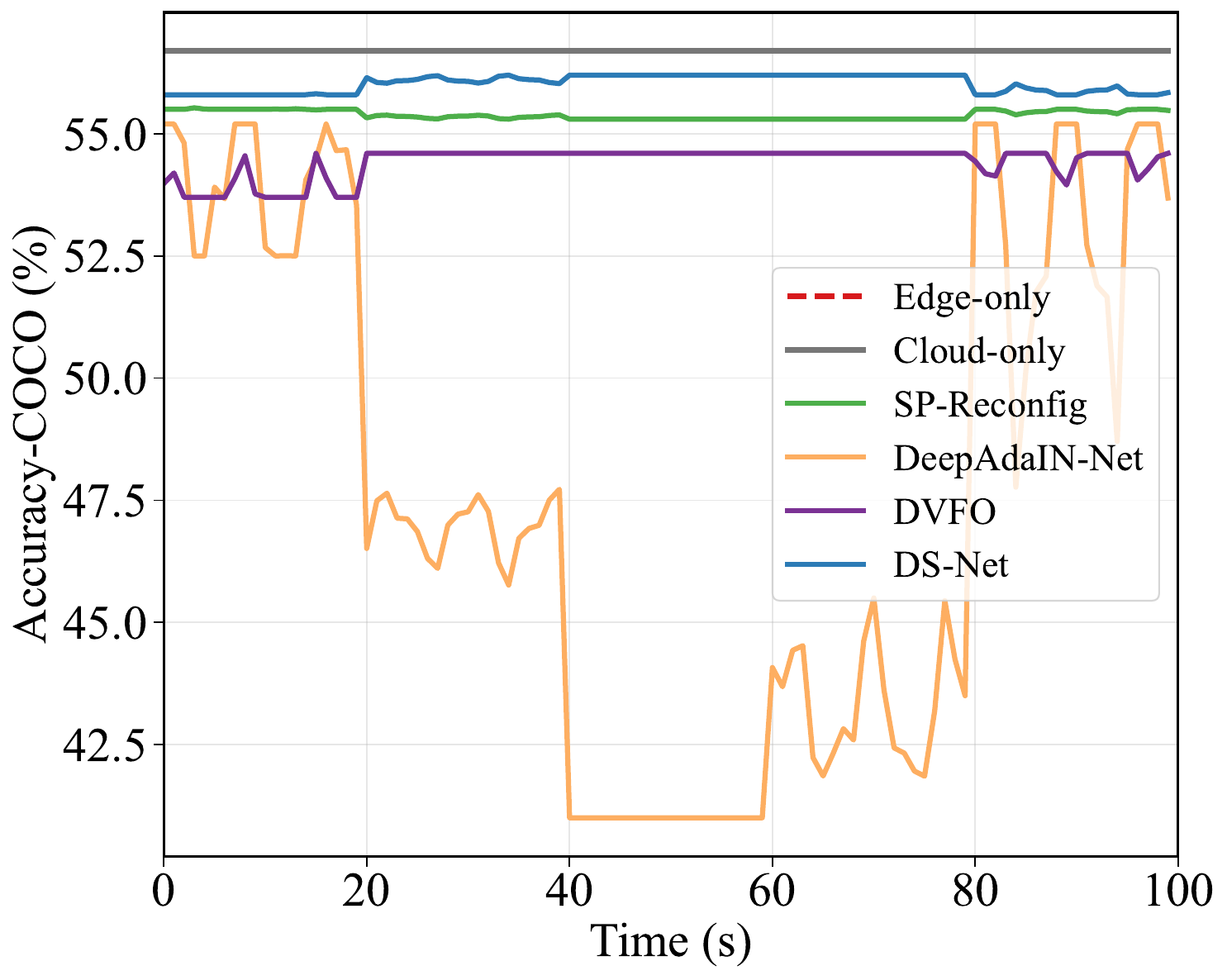}
\end{minipage}%
}
\caption{The robustness of DS-Net is evaluated under non-stationary bandwidth conditions. A fluctuating network environment is constructed by dynamically adjusting the maximum uplink bandwidth over time using \texttt{Wondershaper}. Experimental results show that DS-Net can adaptively adjust split points during bandwidth fluctuations and maintains lower end-to-end latency.}
\label{fig:vbandwidth_tx22}
\end{figure*}



\begin{figure*}[ht]
\centering

\subfigure[Global-profile selection ratio.]{
\begin{minipage}[b]{0.23\textwidth}
\label{fig:sample-adaptive_global_ratio}
\includegraphics[width=1.0\textwidth]{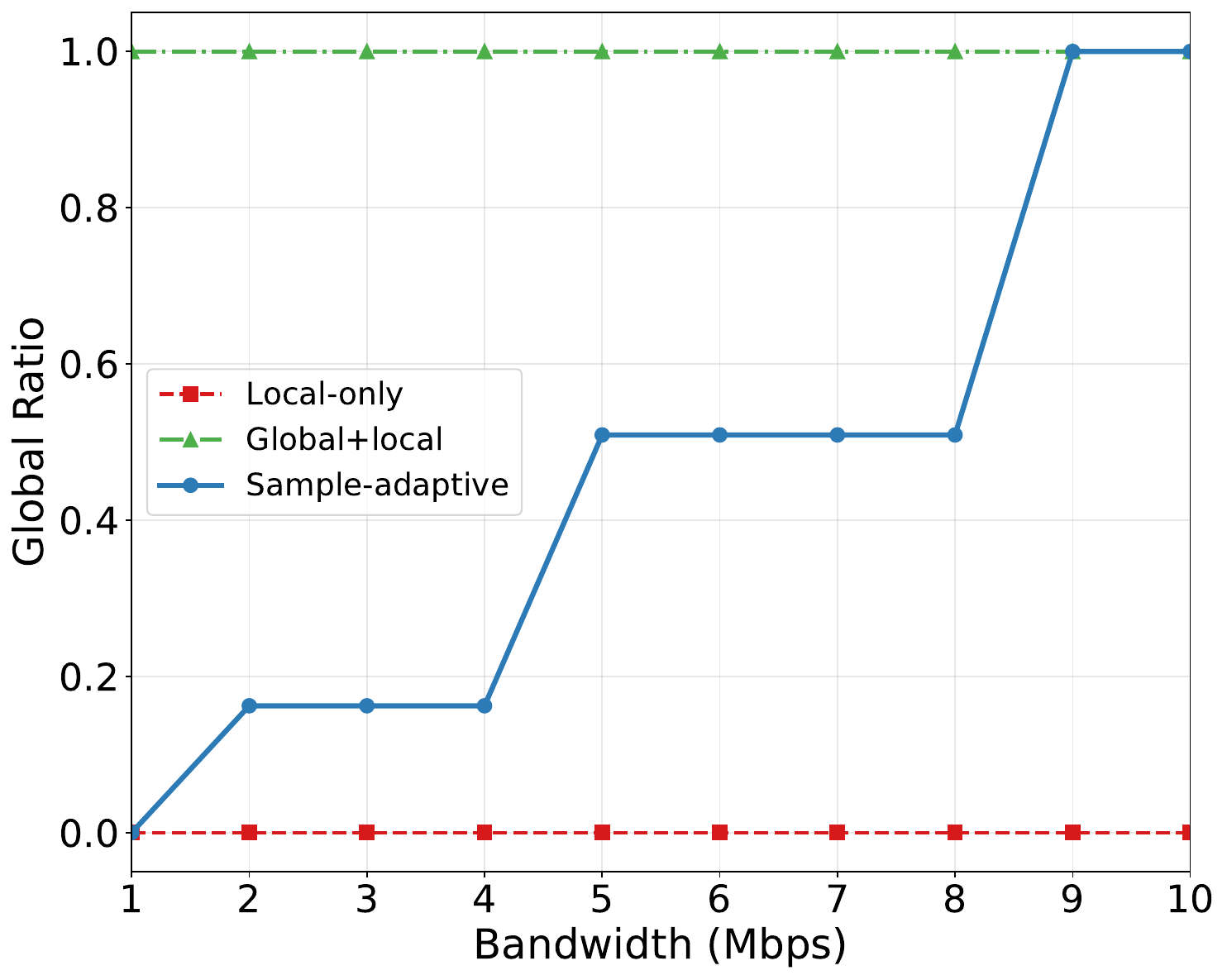}
\end{minipage}%
}
\subfigure[Latency on Nano device.]{
\begin{minipage}[b]{0.23\textwidth}
\label{fig:sample-adaptive_latency}
\includegraphics[width=1.0\textwidth]{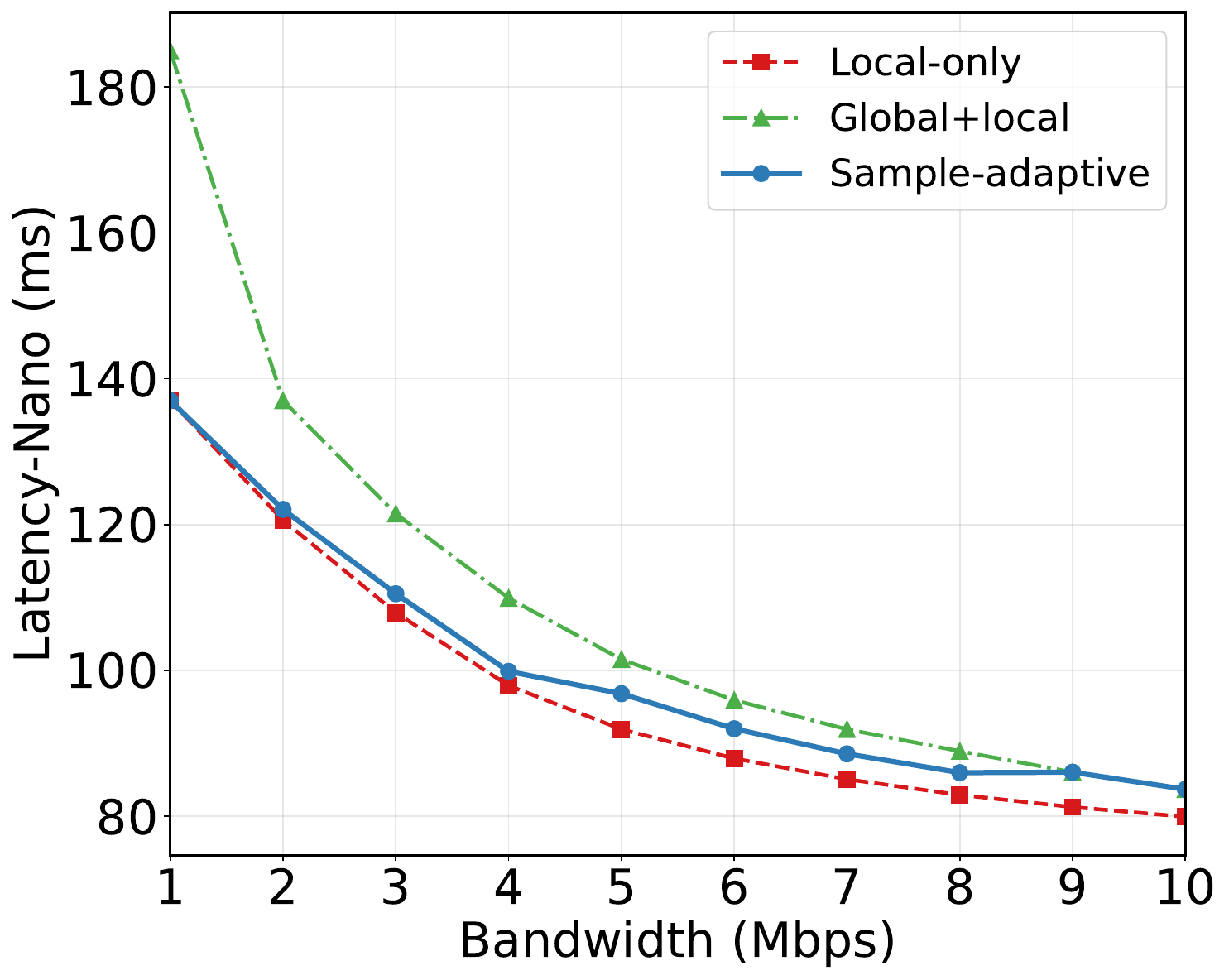}
\end{minipage}%
}
\subfigure[Throughput on Nano device.]{
\begin{minipage}[b]{0.23\textwidth}
\label{fig:sample-adaptive_fps}
\includegraphics[width=1.0\textwidth]{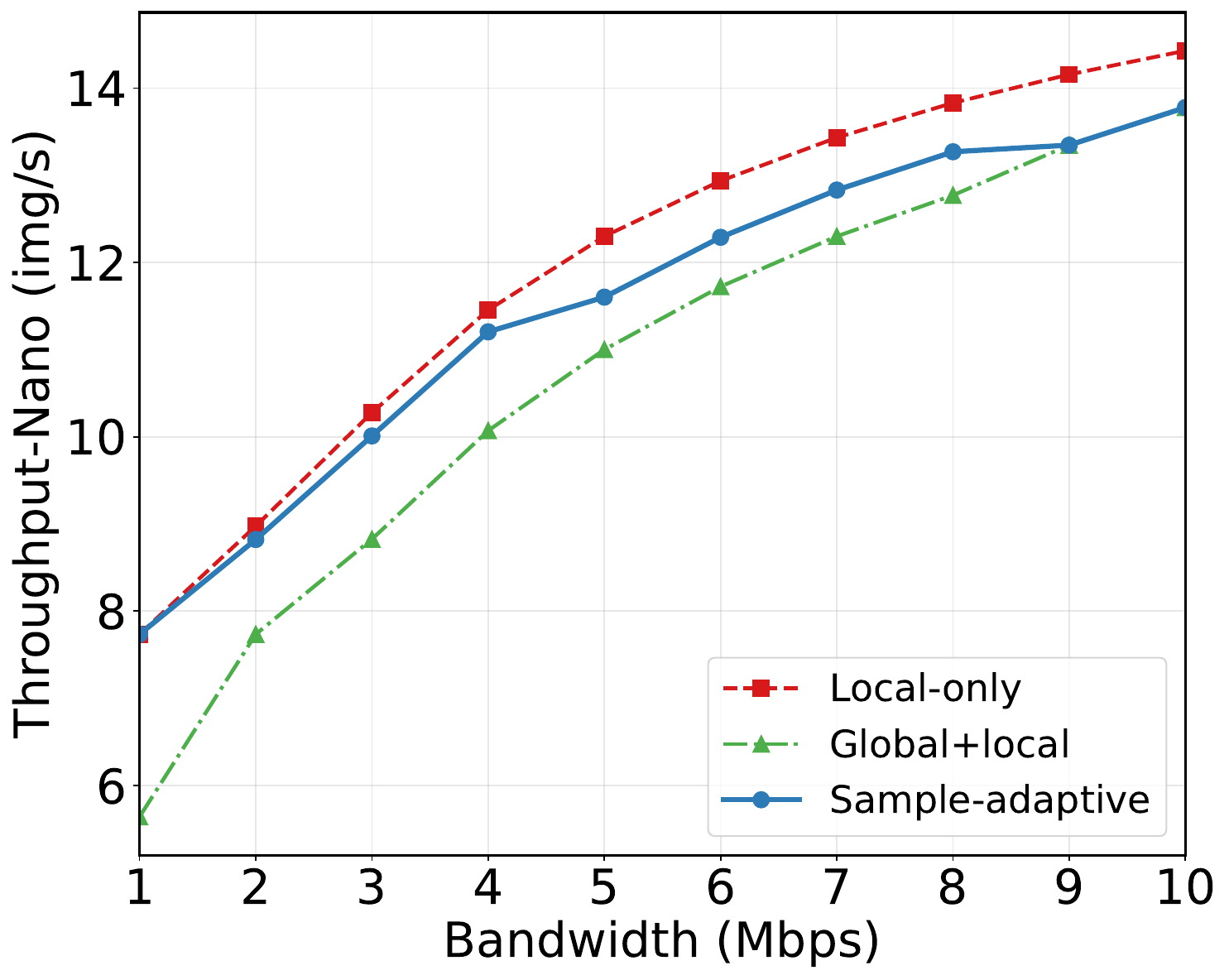}
\end{minipage}%
}
\subfigure[Accuracy on VisDrone dataset.]{
\begin{minipage}[b]{0.23\textwidth}
\label{fig:sample-adaptive_accuracy}
\includegraphics[width=1.0\textwidth]{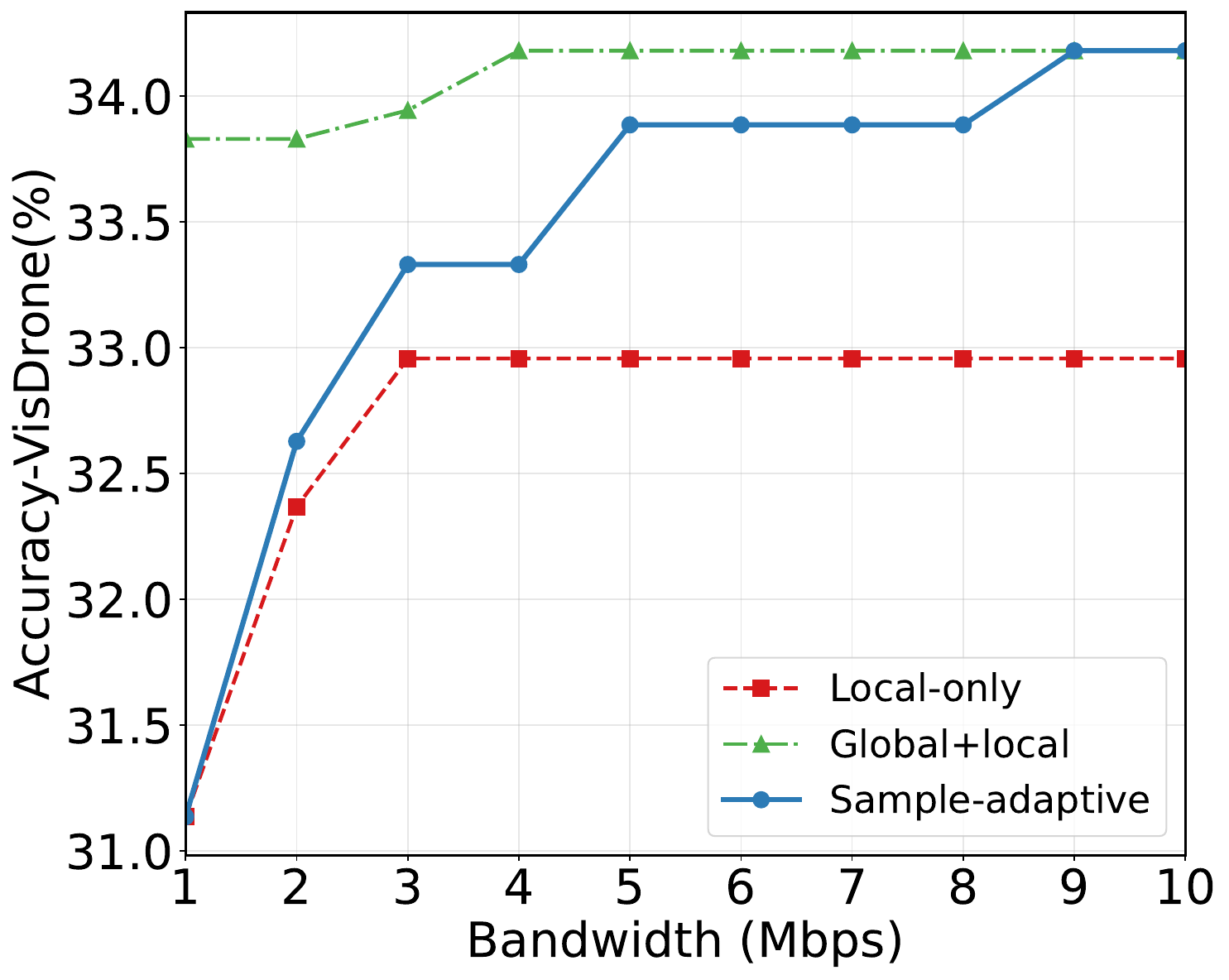}
\end{minipage}%
}
\caption{Performance comparison under different uplink bandwidths on Nano. The proposed sample-adaptive strategy dynamically adjusts the use of global compensation according to the available bandwidth. }
\label{fig:sample-adaptive}
\end{figure*}

\begin{table*}[t]
\centering
\caption{Ablation analysis of the proposed DS-Net based on YOLOv5. Feat. refers to the intermediate feature size at the split layer. The symbol + denotes the addition of a module relative to the method in the preceding row. Notably, layers 5, 7, and 10 are multi-branch structures; through FRN, only single-branch features are preserved in these layers. Channel compress. indicates a 4$\times$ channel reduction, while 1-bit quantization(1-bit quan) achieves approximately 32$\times$ compression ratio.}
\label{tab:abla}
\resizebox{\textwidth}{!}{%
\begin{tabular}{ccccccccc}
\toprule
\multirow{2}{*}{Method} & \multicolumn{2}{c}{Split Layer 3} & \multicolumn{2}{c}{Split Layer 5} & \multicolumn{2}{c}{Split Layer 7} & \multicolumn{2}{c}{Split Layer 10}\\
 & Feat. (KB) & Acc.(\%) (COCO/VisDrone) & Feat. (KB) & Acc.(\%) (COCO/VisDrone) & Feat. (KB) & Acc.(\%) (COCO/VisDrone) & Feat. (KB) & Acc.(\%) (COCO/VisDrone) \\
\hline
Baseline & 3200 & 56.7/35.4 & 4800 & 56.7/35.4 & 5600 & 56.7/35.4 & 5200 & 56.7/35.4 \\
+FRN & 3200 & 55.5/33.4 & 1600 & 55.5/33.4 & 800 & 55.5/33.4 & 400 & 55.5/33.4 \\
+Channel compress. & 800 & 55.4/30.9 & 400 & 56.5/32.1 & 200 & 56.3/31.7 & 100 & 54.6/29.7 \\
+1-bit quan. & 25 & 54.6/30.7 & 12 & 55.2/32.0 & 6 & 54.8/31.5 & 3 & 53.3/29.7 \\
+Local Compen. & 26 & 55.5/32.3 & 15 & 55.6/33.2 & 11 & 55.2/32.3 & 6 & 55.3/31.7 \\
+Global Compen. & 31 & 55.7/33.7 & 18 & 55.9/33.8 & 12 & 55.6/33.5 & 9 & 55.8/33.0 \\
+Local and Global Compen. & \cellcolor{gray!20}\textbf{32} & \cellcolor{gray!20}\textbf{55.8/34.3} & \cellcolor{gray!20}\textbf{21} & \cellcolor{gray!20}\textbf{56.0/34.5} & \cellcolor{gray!20}\textbf{17} & \cellcolor{gray!20}\textbf{56.0/34.5} & \cellcolor{gray!20}\textbf{12} & \cellcolor{gray!20}\textbf{56.2/34.2} \\
\bottomrule
\end{tabular}%
}
\end{table*}







\begin{table}[t]
\centering
\caption{Sensitivity analysis of the loss weight $\lambda_{\mathrm{mask}}$ on YOLOv5. The best results are highlighted in gray.}
\label{tab:loss_weight_sensitivity}
\resizebox{0.65\linewidth}{!}{
\begin{tabular}{c|cc}
\hline
$\lambda_{\mathrm{mask}}$ 
& COCO Acc. (\%) 
& VisDrone Acc. (\%) \\
\hline
0.2 & 55.7 & 33.7 \\
0.4 & 55.8 & 33.5 \\
0.6 & 55.9 & 33.4 \\
0.8 & 56.0 & 33.3 \\
\cellcolor{gray!20}\textbf{1.0} 
& \cellcolor{gray!20}\textbf{56.2} 
& \cellcolor{gray!20}\textbf{34.2} \\
1.2 & 55.9 & 33.6 \\
\hline
\end{tabular}}
\end{table}

\begin{table}[t]
\centering
\caption{Ablation study on predefined and learnable fusion coefficients in global compensation, where the best-performing is highlighted in gray.}
\label{tab:fusion_coefficient_ablation}
\resizebox{\linewidth}{!}{
\begin{tabular}{c|c|cc}
\hline
Fusion Strategy & Coefficients & COCO Acc. (\%) & VisDrone Acc. (\%) \\
\hline
Predefined & $\alpha=0.9,\ \beta=0.1$ & 55.6 & 33.7 \\
Predefined & $\alpha=0.7,\ \beta=0.3$ & 55.4 & 19.0 \\
Predefined & $\alpha=0.6,\ \beta=0.4$ & 55.5 & 33.3 \\
Predefined & $\alpha=0.5,\ \beta=0.5$ & 55.2 & 20.0 \\
\cellcolor{gray!20}\textbf{Learnable}  & \cellcolor{gray!20}\textbf{$\alpha_l=\mathrm{Sigmoid}(\gamma_l),\ \beta_l=1-\alpha_l$} & \cellcolor{gray!20}\textbf{56.2} & \cellcolor{gray!20}\textbf{34.2} \\
\hline
\end{tabular}}
\end{table}

\subsection{Robustness and Adaptability Results of DS-Net}
\noindent\textbf{Robustness under Non-stationary Bandwidth.}
To evaluate DS-Net under non-ideal network conditions, we construct a controlled non-stationary bandwidth environment using \texttt{Wondershaper}. Specifically, the maximum uplink bandwidth between the edge device and the cloud server is adjusted every 20 seconds during a 100-second inference period. The maximum bandwidth is set to 30 Mbps, 10 Mbps, 1 Mbps, 5 Mbps, and 20 Mbps in five consecutive intervals, respectively. Under each maximum bandwidth constraint, the actual available bandwidth still fluctuates during transmission. As shown in Fig.~\ref{fig:vbandwidth2}, the resulting bandwidth trace contains both abrupt bandwidth changes and intra-interval fluctuations. Fig.~\ref{fig:vbandwidth_split2} shows the split-point trajectory of DS-Net. Under low bandwidth conditions (40–80 s), it selects a deeper split point (split 10), to reduce transmission overhead. When sufficient bandwidth is available (80–100 s), it switches to a shallower split point (split 3 or 5) to lower the computational burden on the edge device.

{Fig.~\ref{fig:vbandwidth_lat2} compares the latency performance of different methods under this non-stationary bandwidth environment. During the period of 40-60s when the bandwidth is extremely low, the latency peak of DS-Net is smaller than that of other methods, because it can dynamically select a more appropriate splitting point. In the bandwidth fluctuation state from 20s to 40s, DS-Net can quickly switch between split points 5 and 10 while maintaining a low latency state, which demonstrates that it avoids the reloading overhead for specific split models. Fig.~\ref{fig:vbandwidth_acc2} further shows the accuracy variation over time. DS-Net maintains stable accuracy on the COCO dataset during both steady-state and switching phases, indicating that dynamic split-point adjustment does not introduce significant transient accuracy degradation. Overall, DS-Net achieves a better trade-off between latency stability and accuracy retention under non-stationary bandwidth conditions.

\noindent\textbf{Sample-adaptive Switching Analysis.} 
In the utility function, we set the trade-off coefficient to $\omega=0.22$, which assigns a moderate penalty to latency and encourages the system to switch to lower-latency configurations when network conditions degrade. A larger $\omega$ would make the policy more latency-sensitive and select local compensation more aggressively, while a smaller $\omega$ would favor accuracy and therefore select global compensation more often, especially under acceptable bandwidth.

Fig. ~\ref{fig:sample-adaptive} shows that the proposed sample-adaptive strategy can adjust the compensation mode based on available uplink bandwidth. When the bandwidth is extremely limited, DS-Net avoids using the global path to reduce communication overhead. At a bandwidth of 1 Mbps, the global ratio of this sample-adaptive strategy is 0. As bandwidth increases, DS-Net gradually enables global features for more samples. Within the range of 2–4 Mbps, the global ratio is 0.162, indicating that only a small number of hard samples are assigned to global compensation. At 4 Mbps, DS-Net achieves a latency of 99.85 ms, which is 9.15\% lower than that of the global+local baseline, while the accuracy only drops by 0.85\%. Compared with the local-only baseline, DS-Net improves accuracy by 0.37\% with an additional latency increase of only 1.95 ms. In the medium bandwidth range of 5 to 8 Mbps, the global ratio rises to 0.509, which means that DS-Net selects the global configuration for approximately half of the samples while maintaining the local configuration for the remaining samples, thereby achieving a better trade-off between accuracy and latency. At 8 Mbps, compared with the global+local baseline, DS-Net reduces latency from 88.90 ms to 85.95 ms, representing a 3.31\% latency reduction, a 3.91\% throughput improvement, and only a 0.29\% accuracy decrease. When the bandwidth is sufficiently high, reaching 9–10 Mbps, communication overhead is no longer a critical constraint. Accordingly, DS-Net converges to the global+local configuration, with the global ratio reaching 1.0. Within this range, DS-Net achieves the same accuracy and latency as the global+local baseline, and attains the highest accuracy. Overall, these results demonstrate that instead of statically preferring local-only or global+local execution, DS-Net gradually transitions from local execution to selective global compensation as bandwidth increases, and eventually achieves full global compensation, thereby maintaining a sound balance among latency, throughput and accuracy.

\subsection{Ablation Studies and Visualization Results of DS-Net}
\noindent\textbf{Ablation Analysis.}
Table \ref{tab:abla} summarizes the ablation study results across each candidate split point of the proposed DS-Net, using the YOLOv5 detection model on the COCO and VisDrone datasets. The evaluation includes the output feature volume at each split point and the corresponding detection accuracy. The baseline, which is the original YOLOv5s model without modifications, achieves 56.7\% mAP on COCO and 35.4\% mAP on VisDrone. Notably, layers 5, 7, and 10 are multi-branch feature fusion layers. By integrating the FRN module, redundant inter-branch feature transmission is significantly reduced. For example, at the 10th layer, the transmitted feature size is reduced to 400KB. To further minimize transmission cost, we apply $4\times$ channel reduction, compressing intermediate features from 400KB to 100KB. Subsequent 1-bit quantization reduces data precision from 32 bits to 1 bit, decreasing the feature size to merely 3KB. However, such aggressive compression incurs a noticeable accuracy drop. On COCO, the accuracy degradation is 3.4\%, while on VisDrone, it reaches 5.7\%. This degradation arises because the FRN module has limited capacity to fully reconstruct the original features, and subsequent feature compression intensifies the distribution shift. At split layer 10, adding the local compensation module improves accuracy by about 2\%, while adding the global compensation yields 2.5\% and 3.3\% gains on COCO and VisDrone, respectively. The local compensation focuses on the split layer to help recover the original feature distribution after compression. In contrast, the global compensation not only enhances object awareness but also compensates for the semantic loss caused by feature compression at the edge. As a result, the global compensation provides more complete semantic restoration and achieves greater accuracy improvement than the local one. Finally, compensation modules demonstrate strong performance on the VisDrone dataset, achieving 34.3\% accuracy, with only a 1.1\% drop compared to the original model. On COCO, at the split point 10, the accuracy is only reduced by 0.5\% in the end, and then the features were compressed by 433 times. 

The ablation results in Table X also provide an empirical analysis of error propagation across different split points. Without compensation, aggressive 1-bit quantization leads to clear accuracy degradation at all split layers, indicating that compression errors are propagated through the remaining cloud-side network and further affect FRN-based feature reconstruction. On COCO, the accuracy drops from the baseline 56.7\% to 54.6\%, 55.2\%, 54.8\%, and 53.3\% at split layers 3, 5, 7, and 10, respectively. On VisDrone, the corresponding accuracy decreases from 35.4\% to 30.7\%, 32.0\%, 31.5\%, and 29.7\%, showing that dense small-object scenarios are more sensitive to accumulated reconstruction errors. After introducing both local and global compensation, the final accuracy becomes much more stable across split points. The compensated model achieves 55.8\%, 56.0\%, 56.0\%, and 56.2\% on COCO, and 34.3\%, 34.5\%, 34.5\%, and 34.2\% on VisDrone. These results show that although reconstruction error inevitably exists and may accumulate, the proposed hierarchical compensation effectively bounds its impact. In particular, local compensation mitigates the distribution shift immediately after decoding, while global compensation restores target-aware spatial structures for FRN reconstruction, leading to stable downstream performance across different split configurations.

\noindent\textbf{Analysis of Hyperparameters.} To determine an appropriate weight for the auxiliary mask supervision loss, we evaluate different values of $\lambda$ on COCO and VisDrone. As shown in Table~\ref{tab:loss_weight_sensitivity}, the gray-highlighted row denotes the best-performing setting. When $\lambda=1.0$, DS-Net achieves the highest mAP@0.5 on both COCO and VisDrone, reaching 56.2\% and 34.2\%, respectively. A smaller $\lambda$ may provide insufficient supervision for task-aware mask generation, while an excessively large value may overemphasize the auxiliary mask branch. Therefore, we set $\lambda=1.0$ in the following experiments.

We further compare predefined and learnable fusion coefficients in the global compensation module. As reported in Table~\ref{tab:fusion_coefficient_ablation}, the learnable strategy, highlighted in gray, achieves the best performance on both datasets, with 56.2\% mAP@0.5 on COCO and 34.2\% mAP@0.5 on VisDrone. Compared with fixed coefficient settings, the learnable formulation can adaptively balance the original feature response and the task-aware compensation term for different datasets and feature distributions. In contrast, manually predefined coefficients may be suboptimal and can even cause unstable performance on VisDrone. Therefore, we adopt learnable fusion coefficients in DS-Net.


\subsection{Model Switch Overhead}
Fig.~\ref{fig:switch} illustrates the comparison of storage and memory usage between the traditional independent approach and the DS-Net method across various model architectures. Table \ref{tab:switch} summarizes the switching overhead of both methods on different devices and models. Both approaches support switching among four split points. However, the traditional independent method maintains separate weight files for each split point, leading to significantly higher storage demands. For large-scale models such as ResNet, the independent method requires storing multiple split variants on edge devices, consuming 520 MB of storage space. In contrast, DS-Net adopts a unified architecture that integrates all split points within a single model, thereby reducing storage consumption by 73.3\%. Similarly, for the YOLOv5 model, the independent method occupies 21.2 MB, whereas DS-Net requires merely 11.6 MB, achieving a 45.3\% decrease in storage. In terms of memory usage, the independent method runs multiple parallel processes during inference, resulting in elevated peak memory consumption. On the TX2 platform with the YOLOv8 model, the independent method reaches a peak memory usage of 5124 MB, while DS-Net loads only one model instance, reducing peak memory by 74.6\%. On Nano devices, memory resources are highly constrained, and loading multiple models can easily exceed the available memory capacity. Regarding switching latency, the independent method requires manual switching between split models, which may incur delays exceeding 10 seconds for compute-intensive vision tasks. In contrast, DS-Net reduces this delay to the millisecond level. By leveraging efficient weight sharing, DS-Net minimizes both storage and computational overhead on edge devices while enabling near-instantaneous model switching, making it particularly well-suited for real-time dynamic split point adaptation.

\begin{figure}[t]
\centering
\subfigure[]{   
\begin{minipage}[t]{0.45\linewidth}
\includegraphics[height=4cm,width=4cm]{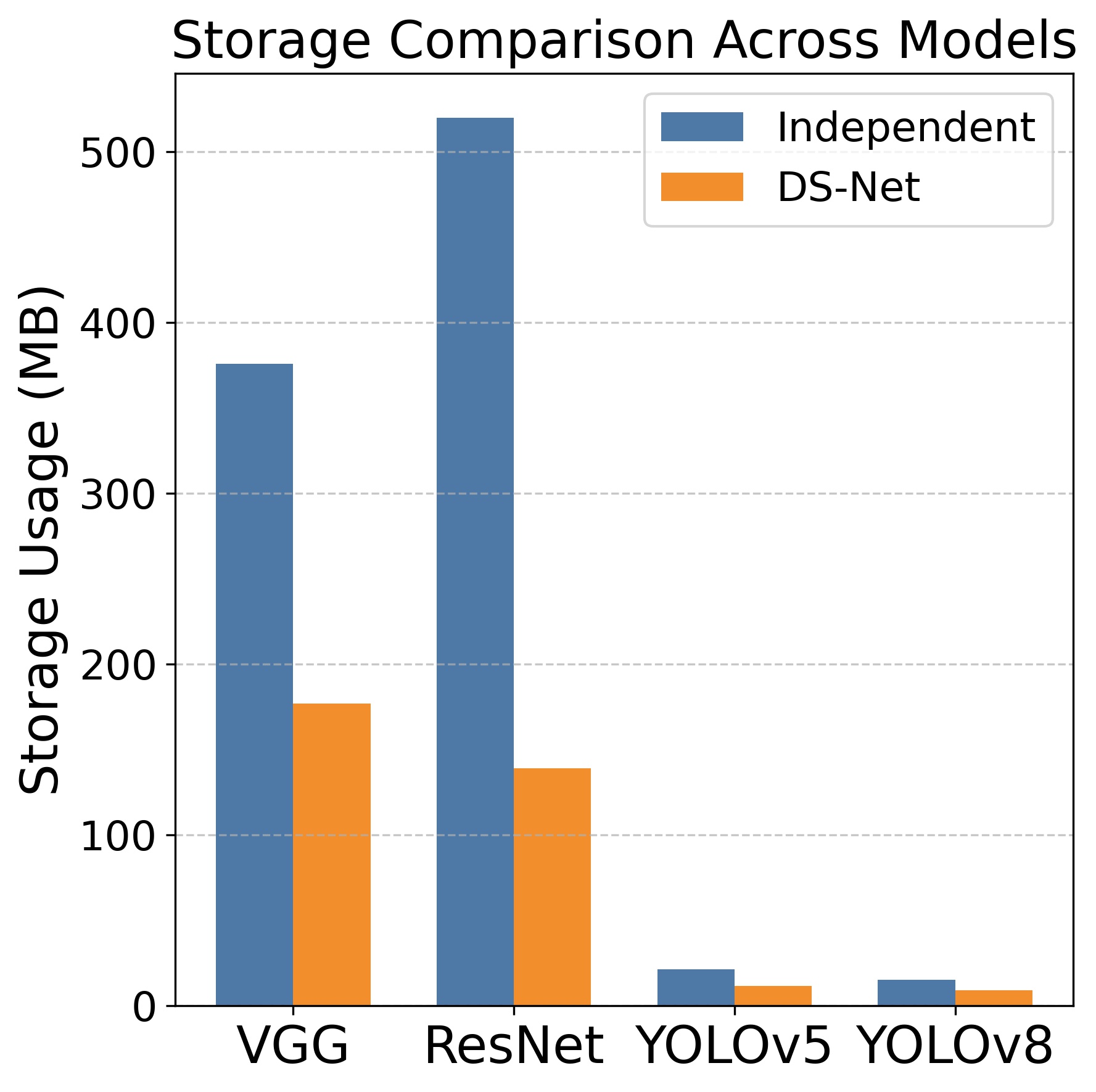}
\end{minipage}%
}
\subfigure[]{
\begin{minipage}[t]{0.48\linewidth}
\includegraphics[height=4cm,width=4cm]{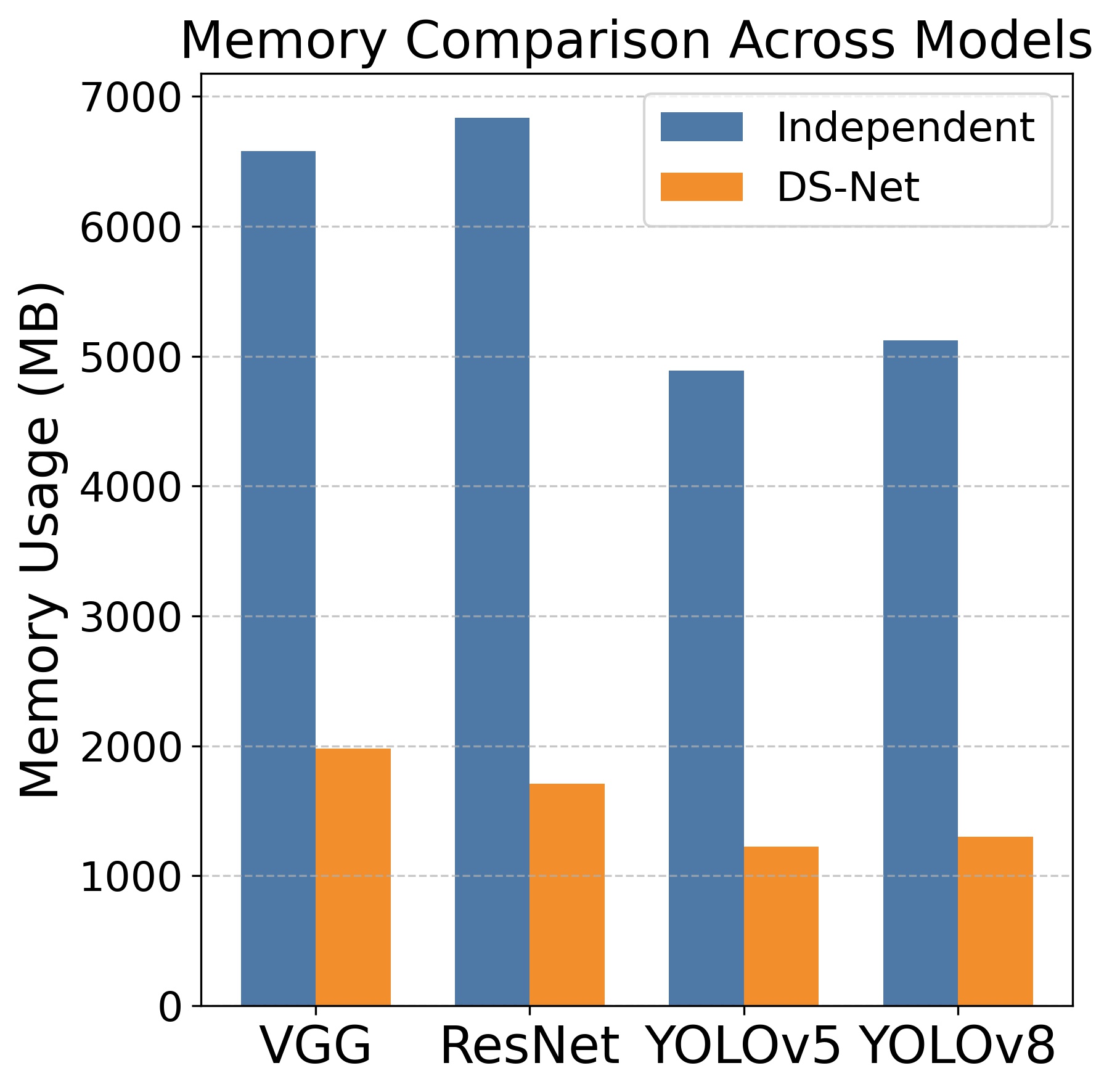}
\end{minipage}%
}
\caption{A comparison of switching latency(in seconds) between the TX2 and Nano platforms across various architectures. Storage denotes the total size of model files stored on the edge device. Memory usage indicates the peak memory consumption during model loading.}    
\label{fig:switch}
\end{figure}

\begin{table}[t]
\centering
\caption{Switching latency(s) comparison on TX2 and Nano across different architectures. Switching latency is defined as the maximum time needed to load and switch between different split point models.}
\label{tab:switch}
\resizebox{\linewidth}{!}{
\begin{tabular}{c c c c c c}
\toprule[1.0pt]
Device & Method & VGG & ResNet & YOLOv5 & YOLOv8 \\
\hline
\multirow{2}{*}{TX2} 
& Independent & $5.60 \times 10^{0}$\,s & $9.60 \times 10^{0}$\,s & $5.70 \times 10^{0}$\,s & $3.50 \times 10^{0}$\,s \\
& \cellcolor{gray!20}DS-Net & \cellcolor{gray!20}\textbf{$3.00 \times 10^{-3}$\,s} & \cellcolor{gray!20}\textbf{$4.00 \times 10^{-3}$\,s} & \cellcolor{gray!20}\textbf{$2.00 \times 10^{-3}$\,s} & \cellcolor{gray!20}\textbf{$5.00 \times 10^{-3}$\,s} \\
\hline
\multirow{2}{*}{Nano} 
& Independent & $1.04 \times 10^{1}$\,s & $1.33 \times 10^{1}$\,s & $6.20 \times 10^{0}$\,s & $3.20 \times 10^{0}$\,s \\
& \cellcolor{gray!20}DS-Net  & \cellcolor{gray!20}\textbf{$2.00 \times 10^{-3}$\,s} & \cellcolor{gray!20}\textbf{$4.00 \times 10^{-3}$\,s} & \cellcolor{gray!20}\textbf{$3.00 \times 10^{-3}$\,s} & \cellcolor{gray!20}\textbf{$4.00 \times 10^{-3}$\,s} \\
\bottomrule
\end{tabular}}
\end{table}

\begin{table}[t]
\centering
\caption{Experimental results for edge-only (TX2), edge-edge (TX-TX2) and multi-edge (TX2-TX2-Nano/TX2) collaborative inference are presented. The bandwidth is set to 20Mbps. Accuracy is evaluated as top-1 accuracy on CIFAR-100 for ResNet and as mAP@0.5 on the VisDrone dataset for YOLOv5.}
\label{tab:mul-split}
\resizebox{\linewidth}{!}{
\begin{tabular}{c|cc c c c c}
\toprule[1.0pt]
Model & Configuration & Split & Thpt. (img/s) & Lat. (ms)  & Acc. (\%)  \\
\hline
\multirow{8}{*}{ResNet} & Edge-only & - & 25 & 39.1 & 78.1 \\
\cline{2-6}
& \multirow{3}{*}{Edge-edge} & 2 & 36 & 42.4 & 78.0 \\
& & 3 & 48 & 40.7 & 77.9 \\
& & 4 & 35 & 39.9 & 77.6 \\
\cline{2-6}
& \multirow{4}{*}{Multi-edge} & (2, 4) & 59 & 40.0 & 74.8 \\
 & & (2, 5) & 43 & 39.8 & 74.2 \\
 & & (3, 4) & 50 & 38.3 & 73.0\\
  & &  (4, 5) & 35 & 37.3 & 73.6\\
\hline
\multirow{9}{*}{YOLOv5} & Edge-only & - & 13 & 77.2 & 35.4 \\
\cline{2-6}
& \multirow{4}{*}{Edge-edge} & 3 & 20 & 90.3 & 34.3 \\
& & 5 & 23 & 85.8 & 34.5 \\
&  & 7 & 21 & 84.1 & 34.5 \\
&  & 10 & 19 & 82.1 & 34.2 \\
\cline{2-6}
& \multirow{4}{*}{Multi-edge} & (3, 5) & 23 & 98.8 & 32.4 \\
 & & (5, 7) & 24 & 92.7 & 33.0 \\
 & & (5, 10) & 24 & 90.7 & 32.7\\
 & &  (7, 10) & 21 & 89.1 & 33.1\\
\bottomrule
\end{tabular}}
\end{table}

\begin{figure}
    \centering
    \includegraphics[width=\linewidth]{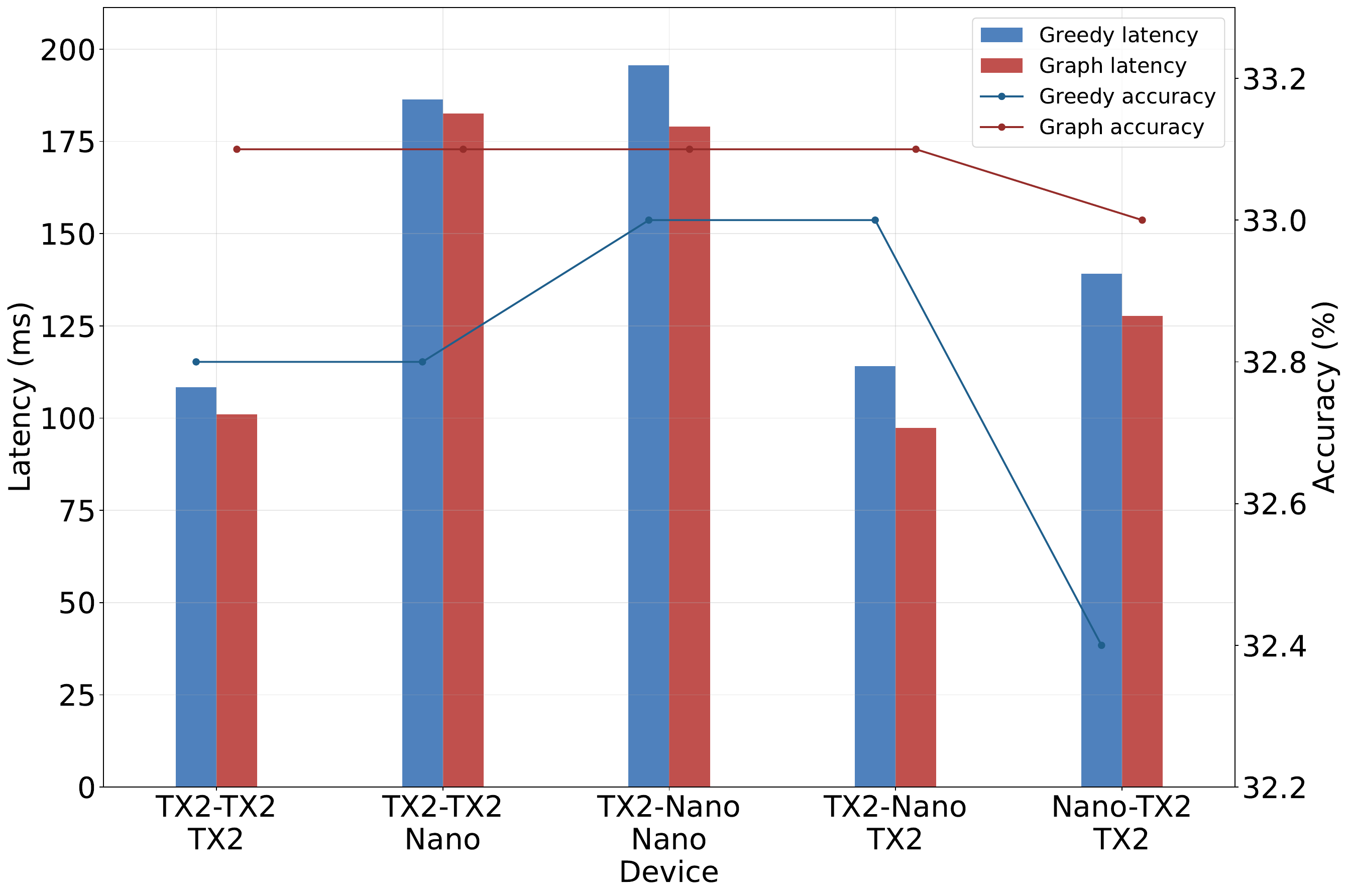}
    \caption{Latency and accuracy comparison of greedy and graph-based multi-split selection under different device chains. Graph-based selection achieves lower latency while preserving stable accuracy.}
\label{fig:graph-geedy}
\end{figure}


\subsection{Multi-Edge Collaborative Expansion and Limitation}


In emergency or adverse environments, communication networks often exhibit instability or even complete disruption, which may make cloud servers inaccessible. To improve the robustness of collaborative inference under such conditions, DS-Net can be extended to a multi-edge serial collaborative inference scenario, where several edge devices jointly execute different segments of the same DNN without relying on cloud computation. In this setting, the model is partitioned across a device chain $d_1 \rightarrow d_2 \rightarrow \cdots \rightarrow d_K$, and each device executes a consecutive network segment before transmitting the intermediate feature to the next device.

To determine the split sequence across multiple heterogeneous devices, we introduce a graph-based joint split selection strategy. Given a candidate split point set $\mathcal{P}$, we jointly select $\mathbf{p}=(p_1,\ldots,p_{K-1})$, where $p_0=0$ and $p_K=N$. Device $d_k$ executes layers from $p_{k-1}+1$ to $p_k$. The objective is to minimize the total computation and transmission latency:
\begin{equation}
\min_{\mathbf{p}}
\sum_{k=1}^{K} T_k^{comp}(p_{k-1},p_k)
+
\sum_{k=1}^{K-1}\frac{S_{p_k}}{B_k},
\label{eq:multi_edge_split}
\end{equation}
where $T_k^{comp}(\cdot)$ denotes the segment computation latency on device $d_k$, $S_{p_k}$ is the transmitted feature size, and $B_k$ is the bandwidth between adjacent devices. This problem is solved by constructing a layered DAG, where each path corresponds to a complete split configuration. The optimal split sequence is obtained by dynamic programming.

Table \ref{tab:mul-split} summarizes the performance of edge-only, edge–edge, and multi-edge inference. For ResNet, the multi-edge split pair (2, 4) achieves a 2.4$\times$ throughput improvement over edge-only inference while incurring only a 3.3\% accuracy loss and maintaining comparable latency. In contrast to edge-edge collaboration, multi-edge collaboration boosts the throughput rate by an additional 1.6$\times$, primarily attributed to the third device offloading computational tasks, enabling the front-end devices to expedite subsequent data processing and thereby enhancing the overall data flow throughput. Benefiting from the proposed global compensation module, the average accuracy degradation across all multi-edge configurations is reduced only 4.2\%. For YOLOv5, the multi-edge split pair (5, 7) achieves a 1.8$\times$ throughput improvement over edge-only inference, with a 2.4\% accuracy drop. Compared with edge–edge inference at split point 3, the multi-edge split pair (3, 5) further improves throughput by 1.2$\times$ with a 1.9\% accuracy decrease. 

As shown in Fig.~\ref{fig:graph-geedy}, the proposed graph-based joint optimization reduces end-to-end latency compared with greedy per-device split selection. The greedy strategy only minimizes the current device computation and current-link transmission cost, which may leave a heavier workload to subsequent devices. In contrast, our method treats the complete multi-device split sequence as a graph path and jointly optimizes computation and transmission costs over the entire device chain. Under 10 Mbps bandwidth, even in the homogeneous TX2-TX2-TX2 setting, the proposed method reduces latency by 6.8\%, indicating that global balancing of layer-wise computation and transmission cost is still necessary even when devices have identical capabilities. The advantage becomes more evident in heterogeneous device chains, where greedy local decisions are more likely to assign unsuitable workloads to weaker devices. For example, the latency is reduced by 14.6\% on TX2-Nano-TX2 and by 8.5\% on TX2-Nano-Nano. Meanwhile, the accuracy remains stable or slightly improved, with gains of 0.3\% on TX2-TX2-TX2 and 0.6\% on Nano-TX2-TX2. This is because the proposed method tends to select globally better split configurations, such as $(7,10)$ or $(5,10)$, rather than the shallow local split configurations selected by the greedy strategy.

\section{Conclusion}\label{sec:clu}
In this paper, we propose DS-Net, a dynamic split network designed for edge-cloud collaborative inference. On the edge side, DS-Net adopts a dynamic switching framework that enables fast and low-overhead switching of split points by adapting to real-time bandwidth and computational capacity. To mitigate the accuracy degradation caused by splitting and compression, a hierarchical compensation module is implemented on the cloud side: local alignment adaptively aligns feature distributions at split points, while global compensation restores structural details to enhance feature reconstruction. We also consider adding sample-adaptive global compensation switch during the process of selecting split points. Experimental results show that DS-Net reduces memory overhead by 74.6\% and achieves millisecond-level switching latency. Compared with state-of-the-art collaborative methods, DS-Net improves accuracy by 1.2\% on the COCO dataset, reduces latency by 42.5\%, and increases throughput by 1.8$\times$ under 10 Mbps on the Nano platform.




\bibliographystyle{IEEEtran}
\bibliography{wx_ref}

\end{document}